\documentclass[10pt,twocolumn,letterpaper]{article}
\usepackage{cvpr}
\usepackage[dvipsnames]{xcolor}
\usepackage[accsupp]{axessibility}
\usepackage{amsmath}
\usepackage{amssymb}
\usepackage{booktabs}
\usepackage{multirow}
\usepackage{xspace}

\definecolor{cvprblue}{rgb}{0.21,0.49,0.74}
\usepackage[pagebackref,breaklinks,colorlinks,citecolor=cvprblue]{hyperref}
\usepackage[capitalize]{cleveref}

\def\paperID{1877}
\def\confName{CVPR}
\def\confYear{2024}

\crefname{section}{Sec.}{Secs.}
\Crefname{section}{Section}{Sections}
\Crefname{table}{Table}{Tables}
\crefname{table}{Tab.}{Tabs.}

\newcommand{\method}[0]{3D-Fauna\xspace}

\makeatletter
\renewcommand{\paragraph}{%
  \@startsection{paragraph}{4}%
  {\z@}{-0.5em}{-0.5em}%
  {\normalfont\normalsize\bfseries}%
}
\newcommand{\printfnsymbol}[1]{%
        \textsuperscript{\@fnsymbol{#1}}%
}
\makeatother

\newcommand\rurl[1]{%
  \href{https://#1}{\nolinkurl{#1}}%
}
\newif\ifarxiv
\arxivtrue %

\title{Learning the 3D Fauna of the Web}

\author{
Zizhang Li\textsuperscript{1}\printfnsymbol{1} \; 
Dor Litvak\textsuperscript{1,2}\printfnsymbol{1} \; 
Ruining Li\textsuperscript{3} \; 
Yunzhi Zhang\textsuperscript{1} \; 
Tomas Jakab\textsuperscript{3} \; 
Christian Rupprecht\textsuperscript{3} \; 
\\[0.2em]
Shangzhe Wu\textsuperscript{1\dag} \; 
Andrea Vedaldi\textsuperscript{3\dag} \; 
Jiajun Wu\textsuperscript{1\dag} \; 
\\[0.5em]
\textsuperscript{1}Stanford University \quad
\textsuperscript{2}UT Austin \quad
\textsuperscript{3}University of Oxford
\\
{\small\rurl{kyleleey.github.io/3DFauna/}}
}

\begin{document}

\twocolumn[\maketitle\vspace{-3em}\begin{center}
    \includegraphics[trim={5px 10px 5px 15px}, clip, width=\linewidth]{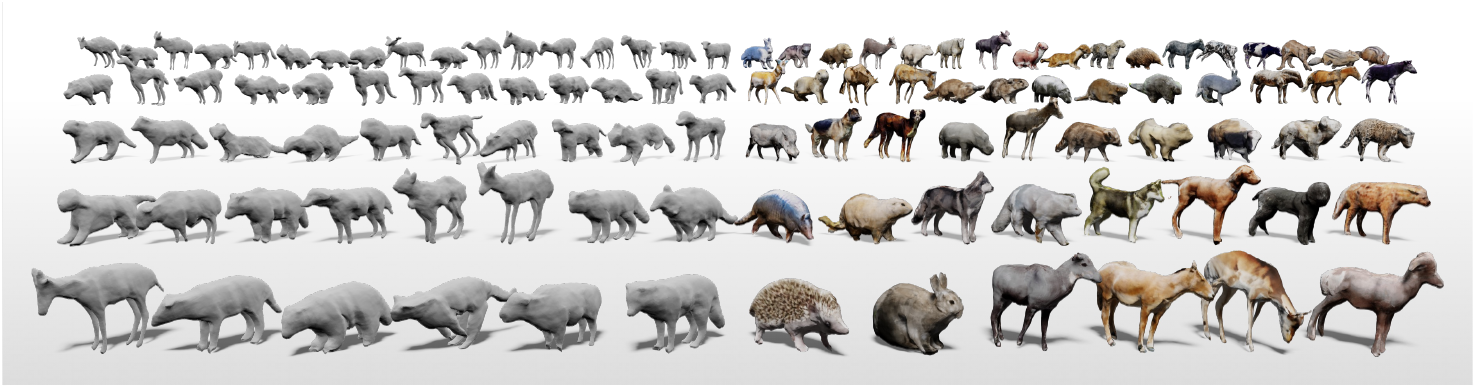}
\end{center}\vspace{-1.5em}
\captionof{figure}{\textbf{Learning Diverse 3D Animals from the Internet.}
Our method, \emph{\method}, learns a pan-category deformable 3D model of more than 100 different animal species
using only 2D Internet images as training data.
At test time, the model can turn a single image of an quadruped instance into an articulated, textured 3D mesh in a feed-forward manner, ready for animation and rendering.
}%
\label{fig:teaser}
\bigbreak]

\def\thefootnote{*}\footnotetext{Equal contribution}\def\thefootnote{\arabic{footnote}}
\def\thefootnote{\dag}\footnotetext{Equal advising}\def\thefootnote{\arabic{footnote}}

\newcommand{\netname}{\textbf{RICO}}

\newcommand{\ba}{\boldsymbol{a}}\newcommand{\bA}{\boldsymbol{A}}
\newcommand{\bb}{\boldsymbol{b}}\newcommand{\bB}{\boldsymbol{B}}
\newcommand{\bc}{\boldsymbol{c}}\newcommand{\bC}{\boldsymbol{C}}
\newcommand{\bd}{\boldsymbol{d}}\newcommand{\bD}{\boldsymbol{D}}
\newcommand{\be}{\boldsymbol{e}}\newcommand{\bE}{\boldsymbol{E}}
\newcommand{\bff}{\boldsymbol{f}}\newcommand{\bF}{\boldsymbol{F}} %
\newcommand{\bg}{\boldsymbol{g}}\newcommand{\bG}{\boldsymbol{G}}
\newcommand{\bh}{\boldsymbol{h}}\newcommand{\bH}{\boldsymbol{H}}
\newcommand{\bi}{\boldsymbol{i}}\newcommand{\bI}{\boldsymbol{I}}
\newcommand{\bj}{\boldsymbol{j}}\newcommand{\bJ}{\boldsymbol{J}}
\newcommand{\bk}{\boldsymbol{k}}\newcommand{\bK}{\boldsymbol{K}}
\newcommand{\bl}{\boldsymbol{l}}\newcommand{\bL}{\boldsymbol{L}}
\newcommand{\bm}{\boldsymbol{m}}\newcommand{\bM}{\boldsymbol{M}}
\newcommand{\bn}{\boldsymbol{n}}\newcommand{\bN}{\boldsymbol{N}}
\newcommand{\bo}{\boldsymbol{o}}\newcommand{\bO}{\boldsymbol{O}}
\newcommand{\bp}{\boldsymbol{p}}\newcommand{\bP}{\boldsymbol{P}}
\newcommand{\bq}{\boldsymbol{q}}\newcommand{\bQ}{\boldsymbol{Q}}
\newcommand{\br}{\boldsymbol{r}}\newcommand{\bR}{\boldsymbol{R}}
\newcommand{\bs}{\boldsymbol{s}}\newcommand{\bS}{\boldsymbol{S}}
\newcommand{\bt}{\boldsymbol{t}}\newcommand{\bT}{\boldsymbol{T}}
\newcommand{\bu}{\boldsymbol{u}}\newcommand{\bU}{\boldsymbol{U}}
\newcommand{\bv}{\boldsymbol{v}}\newcommand{\bV}{\boldsymbol{V}}
\newcommand{\bw}{\boldsymbol{w}}\newcommand{\bW}{\boldsymbol{W}}
\newcommand{\bx}{\boldsymbol{x}}\newcommand{\bX}{\boldsymbol{X}}
\newcommand{\by}{\boldsymbol{y}}\newcommand{\bY}{\boldsymbol{Y}}
\newcommand{\bz}{\boldsymbol{z}}\newcommand{\bZ}{\boldsymbol{Z}}

\newcommand{\balpha}{\boldsymbol{\alpha}}\newcommand{\bAlpha}{\boldsymbol{\Alpha}}
\newcommand{\bbeta}{\boldsymbol{\beta}}\newcommand{\bBeta}{\boldsymbol{\Beta}}
\newcommand{\bgamma}{\boldsymbol{\gamma}}\newcommand{\bGamma}{\boldsymbol{\Gamma}}
\newcommand{\bdelta}{\boldsymbol{\delta}}\newcommand{\bDelta}{\boldsymbol{\Delta}}
\newcommand{\bepsilon}{\boldsymbol{\epsilon}}\newcommand{\bEpsilon}{\boldsymbol{\Epsilon}}
\newcommand{\bzeta}{\boldsymbol{\zeta}}\newcommand{\bZeta}{\boldsymbol{\Zeta}}
\newcommand{\beeta}{\boldsymbol{\eta}}\newcommand{\bEta}{\boldsymbol{\Eta}} %
\newcommand{\btheta}{\boldsymbol{\theta}}\newcommand{\bTheta}{\boldsymbol{\Theta}}
\newcommand{\biota}{\boldsymbol{\iota}}\newcommand{\bIota}{\boldsymbol{\Iota}}
\newcommand{\bkappa}{\boldsymbol{\kappa}}\newcommand{\bKappa}{\boldsymbol{\Kappa}}
\newcommand{\blambda}{\boldsymbol{\lambda}}\newcommand{\bLambda}{\boldsymbol{\Lambda}}
\newcommand{\bmu}{\boldsymbol{\mu}}\newcommand{\bMu}{\boldsymbol{\Mu}}
\newcommand{\bnu}{\boldsymbol{\nu}}\newcommand{\bNu}{\boldsymbol{\Nu}}
\newcommand{\bxi}{\boldsymbol{\xi}}\newcommand{\bXi}{\boldsymbol{\Xi}}
\newcommand{\bomikron}{\boldsymbol{\omikron}}\newcommand{\bOmikron}{\boldsymbol{\Omikron}}
\newcommand{\bpi}{\boldsymbol{\pi}}\newcommand{\bPi}{\boldsymbol{\Pi}}
\newcommand{\brho}{\boldsymbol{\rho}}\newcommand{\bRho}{\boldsymbol{\Rho}}
\newcommand{\bsigma}{\boldsymbol{\sigma}}\newcommand{\bSigma}{\boldsymbol{\Sigma}}
\newcommand{\btau}{\boldsymbol{\tau}}\newcommand{\bTau}{\boldsymbol{\Tau}}
\newcommand{\bypsilon}{\boldsymbol{\ypsilon}}\newcommand{\bYpsilon}{\boldsymbol{\Ypsilon}}
\newcommand{\bphi}{\boldsymbol{\phi}}\newcommand{\bPhi}{\boldsymbol{\Phi}}
\newcommand{\bchi}{\boldsymbol{\chi}}\newcommand{\bChi}{\boldsymbol{\Chi}}
\newcommand{\bpsi}{\boldsymbol{\psi}}\newcommand{\bPsi}{\boldsymbol{\Psi}}
\newcommand{\bomega}{\boldsymbol{\omega}}\newcommand{\bOmega}{\boldsymbol{\Omega}}

\newcommand{\nA}{\mathbb{A}}
\newcommand{\nB}{\mathbb{B}}
\newcommand{\nC}{\mathbb{C}}
\newcommand{\nD}{\mathbb{D}}
\newcommand{\nE}{\mathbb{E}}
\newcommand{\nF}{\mathbb{F}}
\newcommand{\nG}{\mathbb{G}}
\newcommand{\nH}{\mathbb{H}}
\newcommand{\nI}{\mathbb{I}}
\newcommand{\nJ}{\mathbb{J}}
\newcommand{\nK}{\mathbb{K}}
\newcommand{\nL}{\mathbb{L}}
\newcommand{\nM}{\mathbb{M}}
\newcommand{\nN}{\mathbb{N}}
\newcommand{\nO}{\mathbb{O}}
\newcommand{\nP}{\mathbb{P}}
\newcommand{\nQ}{\mathbb{Q}}
\newcommand{\nR}{\mathbb{R}}
\newcommand{\nS}{\mathbb{S}}
\newcommand{\nT}{\mathbb{T}}
\newcommand{\nU}{\mathbb{U}}
\newcommand{\nV}{\mathbb{V}}
\newcommand{\nW}{\mathbb{W}}
\newcommand{\nX}{\mathbb{X}}
\newcommand{\nY}{\mathbb{Y}}
\newcommand{\nZ}{\mathbb{Z}}

\newcommand{\cA}{\mathcal{A}}
\newcommand{\cB}{\mathcal{B}}
\newcommand{\cC}{\mathcal{C}}
\newcommand{\cD}{\mathcal{D}}
\newcommand{\cE}{\mathcal{E}}
\newcommand{\cF}{\mathcal{F}}
\newcommand{\cG}{\mathcal{G}}
\newcommand{\cH}{\mathcal{H}}
\newcommand{\cI}{\mathcal{I}}
\newcommand{\cJ}{\mathcal{J}}
\newcommand{\cK}{\mathcal{K}}
\newcommand{\cL}{\mathcal{L}}
\newcommand{\cM}{\mathcal{M}}
\newcommand{\cN}{\mathcal{N}}
\newcommand{\cO}{\mathcal{O}}
\newcommand{\cP}{\mathcal{P}}
\newcommand{\cQ}{\mathcal{Q}}
\newcommand{\cR}{\mathcal{R}}
\newcommand{\cS}{\mathcal{S}}
\newcommand{\cT}{\mathcal{T}}
\newcommand{\cU}{\mathcal{U}}
\newcommand{\cV}{\mathcal{V}}
\newcommand{\cW}{\mathcal{W}}
\newcommand{\cX}{\mathcal{X}}
\newcommand{\cY}{\mathcal{Y}}
\newcommand{\cZ}{\mathcal{Z}}

\newcommand{\figref}[1]{Fig.~\ref{#1}}
\newcommand{\secref}[1]{Section~\ref{#1}}
\newcommand{\algref}[1]{Algorithm~\ref{#1}}
\newcommand{\eqnref}[1]{Eq.~\eqref{#1}}
\newcommand{\tabref}[1]{Table~\ref{#1}}

\def\mc{\mathcal}
\def\mb{\boldsymbol}

\newcommand{\T}{^{\raisemath{-1pt}{\mathsf{T}}}}

\newcommand{\Perp}{\perp\!\!\! \perp}

\makeatletter
\DeclareRobustCommand\onedot{\futurelet\@let@token\@onedot}
\def\@onedot{\ifx\@let@token.\else.\null\fi\xspace}
\def\eg{e.g\onedot} \def\Eg{E.g\onedot}
\def\ie{i.e\onedot} \def\Ie{I.e\onedot}
\def\cf{cf\onedot} \def\Cf{Cf\onedot}
\def\etc{etc\onedot}
\def\vs{vs\onedot}
\def\wrt{wrt\onedot}
\def\dof{d.o.f\onedot}
\def\etal{et~al\onedot}
\def\iid{i.i.d\onedot}
\def\evs{\emph{vs}\onedot}
\makeatother

\newcommand*\rot{\rotatebox{90}}

\newcommand{\boldparagraph}[1]{\vspace{0.4em}\noindent{\bf #1:}}

\definecolor{darkgreen}{rgb}{0,0.7,0}
\definecolor{lightred}{rgb}{1.,0.5,0.5}

\begin{abstract}
Learning 3D models of all animals in nature requires massively scaling up existing solutions. With this ultimate goal in mind, we develop 3D-Fauna, an approach that learns a pan-category deformable 3D animal model for more than 100 animal species jointly.
One crucial bottleneck of modeling animals is the limited availability of training data, which we overcome by learning our model from 2D Internet images.
We show that prior approaches, which are category-specific, fail to generalize to rare species with limited training images.
We address this challenge by introducing the Semantic Bank of Skinned Models (SBSM), which automatically discovers a small set of base animal shapes by combining geometric inductive priors with semantic knowledge implicitly captured by an off-the-shelf self-supervised feature extractor.
To train such a model, we also contribute a new large-scale dataset of diverse animal species.
At inference time, given a single image of any quadruped animal, our model reconstructs an articulated 3D mesh in a feed-forward manner in seconds.
\vspace{-1em}
\end{abstract}
\section{Introduction}%
\label{s:intro}

Computer vision models can nowadays reconstruct humans in monocular images and videos robustly and accurately, recovering their 3D shape, articulated pose, and even appearance~\cite{fischler73the-representation,felzenszwalb2000efficient,loper2015smpl,Bogo16,joo2019panoptic,goel2023humans}.
However, humans are but a tiny fraction of the animals that exist in nature, and 3D models remain essentially blind to the vast majority of biodiversity.

While in principle the same approaches that work for humans could work for many other animal species, in practice scaling it to each of the 2.1 million different animal species on Earth is nearly hopeless.
In fact, building a human model such as SMPL~\cite{loper2015smpl} and a corresponding pose predictor~\cite{Bogo16,goel2023humans} requires collecting 3D scans of many people in laboratory~\cite{joo2019panoptic}, crafting a corresponding articulated deformable model semi-automatically, and collecting extensive manual labels to train corresponding pose regressors.
Of all animals, only humans are currently of sufficient importance in applications to justify the costs.

A technically harder but much more practical approach is to learn animal models automatically from images and videos readily available on the Internet.
Several authors have demonstrated that at least rough models can be learned from such uncontrolled image collections~\cite{kanazawa18cmr, wu2023magicpony, yao2022lassie}.
Even so, many limitations remain, starting from the fact that these methods can only reconstruct one or a few specific animal exemplars~\cite{yao2022lassie}, or at most a single class of animals at a given time~\cite{kanazawa18cmr,wu2023magicpony}.
The latter restriction is particularly glaring, as it defeats the purpose of using the Internet as a vast data source for modeling biodiversity.

We introduce \emph{\method}, a method that learns a pan-category deformable model for a large number ($>100$) of different quadruped animal species, such as dogs, antelopes, and hedgehogs, as shown in \cref{fig:teaser}.
For the approach to be as automated and thus as scalable as possible, we assume that \emph{only} Internet images of the animals are provided as training data and only consider as prerequisites a pre-trained 2D object segmentation model and off-the-shelf unsupervised visual features.
\method is designed as a feed-forward network that deforms and poses the deformable model to reconstruct any animal given a single image as input.
The ability to perform monocular reconstruction is necessary for training on (single-view) Internet images, and is also useful in many real-world applications.

Crucial to \method is to learn a \emph{single joint model} of \emph{all animals} in one go.
Despite posing a challenge, modeling many animals jointly is essential for reconstructing rarer species, for which we often have only a small number of images to train on.
This allows us to exploit the structural similarity of different animals that results from evolution, and maximize statistical efficiency.
Here, we focus our attention on animals that share a given body plan, in particular, quadrupeds, and share the structure of the underlying skeletal model, which would otherwise be difficult to pin down.

Learning such a model from only unlabeled single-view images requires several technical innovations.
The most important is to develop a 3D representation that is sufficiently \emph{expressive} to model the diverse shape variations of the animals, and at the same time \emph{tight} enough to be learned from single-view images without overfitting individual views.
Prior work partly achieved this goal by using skinned models, which consider small shape variations around a base template followed by articulation~\cite{wu2023magicpony}.
We found that this approach does not provide sufficient inductive biases to learn \emph{diverse} animal species from Internet images alone.
Hence, we introduce the \emph{Semantic Bank of Skinned Models} (SBSM), which uses off-the-shelf unsupervised features, such as DINO~\cite{caron2021emerging,oquab2023dinov2}, to hypothesize how different animals may relate semantically, and automatically learns a low-dimensional base shape bank.

Lastly, Internet images, which are not captured with the purpose of 3D reconstruction in mind, are characterized by a strong photographer bias, skewing the viewpoint distribution to mostly frontal, which significantly hinders the stability of 3D shape learning.
To mitigate this issue, \method further encourages the predicted shapes to look realistic from all viewpoints, by introducing an efficient mask discriminator that enforces the silhouettes rendered from a \emph{random} viewpoint to stay within the distribution of the silhouettes of the real images.

Combining these ideas, \method is an end-to-end framework that learns a pan-category model of 3D quadruped animals from online image collections.
To train \method, we collected a large-scale animal dataset of over 100 quadruped species, dubbed the \emph{Fauna Dataset}, as part of the contribution.
After training, the model can turn a single test image of any quadruped instance into a fully articulated 3D mesh in a feed-forward fashion, ready for animation and rendering.
Extensive quantitative and qualitative comparisons demonstrate significant improvements over existing methods.
Code and data will be released.

\section{Related Work}%
\label{s:related}

\begin{figure*}[t]
\begin{center}
\includegraphics[trim={0, 20, 0px, 0}, width=\linewidth]{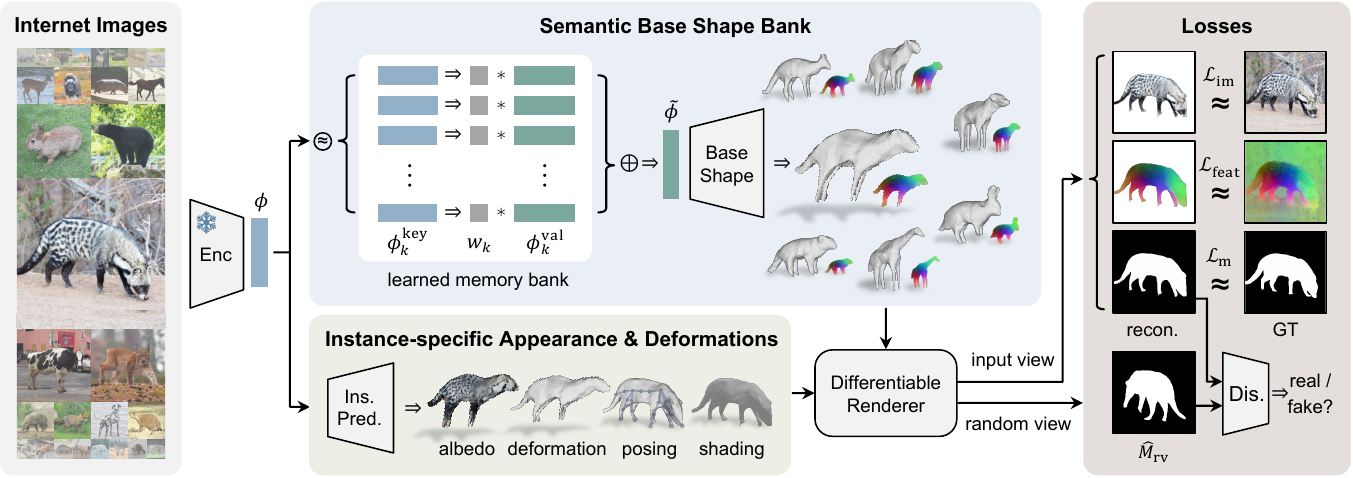}
\end{center}
   \caption{\textbf{Training Pipeline.}
   \method is trained using only single-view images from the Internet.
   Given each input image, it first extracts a feature vector $\phi$ using a pre-trained unsupervised image encoder~\cite{caron2021emerging}.
   This is then used to query a learned memory bank to produce a base shape and a DINO feature field in the canonical pose.
   The model also predicts the albedo, instance-specific deformation, articulated pose and lighting, and is trained via image reconstruction losses on RGB, DINO feature map and mask, as well as a mask discriminator loss.
   }
   \vspace{-0.1in}
\label{fig:overview}
\end{figure*}

\paragraph{Optimization-Based 3D Reconstruction of Animals.}

Due to the lack of explicit 3D data for the vast majority of animals, reconstruction has mostly relied on pre-defined shape models or multi-view images.
Initially, efforts focus on fitting a parametric 3D shape model obtained form 3D scans, \eg, SMAL~\cite{zuffi20173d}, to animal images using annotated 2D keypoints and segmentation masks, which is further extended to multi-view images~\cite{zuffi2018lions}.
Other works aim to optimize the 3D shape~\cite{Cashman12dolphin, wang2021birds, yang2021viser, yang2022banmo, yao2022lassie, yao2023hi, yang2023rac, yao2023artic3d} directly from image or video collections of a smaller scale using various forms of supervision in addition to masks, such as keypoints~\cite{Cashman12dolphin, wang2021birds}, self-supervised semantic correspondences~\cite{yao2022lassie, yao2023hi, yao2023artic3d}, optical flow~\cite{yang21lasr, yang2021viser, yang2022banmo, yang2023rac}, surface normals~\cite{yang2023rac}, category-specific template shapes~\cite{Cashman12dolphin, wang2021birds}.

\paragraph{Learning 3D from Internet Images and Videos.}%
\label{related:internet}

Recently, authors have attempted to learn 3D priors from Internet images and videos at a larger scale~\cite{tulsiani2020imr, wu2020unsupervised, wu2021rendering, Goel2020ucmr, kulkarni2020articulation, li2020self, ye21shelf, alwala2022pre, kanazawa18cmr, wu2023dove, wu2023magicpony, jakab2023farm3d}, mostly focusing on a single category at a time.
Reconstructing animals presents additional challenges due to their highly deformable nature, which often necessitates stronger supervisory signals for training, similar to the ones used in optimization-based methods.
Some methods have, in particular, learned to model articulated animals, such as horses, from single-view image collections without any 3D supervision, adopting a hierarchical shape model that factorizes a category-specific prior shape from instance-specific shape deformation and articulation~\cite{wu2023dove, wu2023magicpony, jakab2023farm3d}.
However, these models are trained in a category-specific manner and fail to generalize to less common animal species as shown in \cref{sec:qual_comparison}.

Attempts to model diverse animal species again resort to pre-defined shape models, \eg, SMAL\@.
Ruegg~\etal~\cite{ruegg2022barc, ruegg2023bite} model multiple dog breeds and regularize the learning process by encouraging intra-breed similarities using a triplet loss, which requires breed labels for training, in addition to keypoint annotations and template shape models.
In contrast, our approach reconstructs a significantly broader set of animals and is trained in a category-agnostic fashion, without relying on existing 3D shape models or keypoints.
Another related work~\cite{huang2023shapeclipper} aims to learn a category-agnostic 3D shape regressor by exploiting pre-trained CLIP features and an off-the-shelf normal estimator,
but does not model deformation and produces coarse shapes.
Concurrent work SAOR~\cite{aygun2023saor} also trains one model to reconstruct diverse animal categories, but obtains less realistic results and tends to suffer from strong photographer bias.

Another line of research attempts to distill 3D reconstructions from 2D generative models trained on large-scale datasets of Internet images, which can be GAN-based~\cite{goodfellow2014generative, nguyen2019hologan, chan2021piGAN, Chan2022} or more recently, diffusion-based models~\cite{ho2020denoising, song2020score, melas2023realfusion, deng2022nerdi} using Score Distillation Sampling~\cite{poole2022dreamfusion} and its variants.
This idea has been extended to learn image-conditional multi-view generator networks~\cite{liu2023zero,qian2023magic123,liu2023one,dmv2023,sun2023dreamcraft3d,long2023wonder3d,yang2023consistnet,weng2023consistent123,tang2023dreamgaussian,liu2023syncdreamer,shi2023mvdream,kim2023collaborative}.
However, most of these methods optimize one single shape at a time, whereas our model learns a pan-category deformable model that can reconstruct any animal instance in a feed-forward fashion.

\paragraph{Animal Datasets.}%
\label{related:animal datasets}

Learning 3D models often requires high-quality images without blur or occlusion.
Existing high-quality datasets were only collected for a small number of categories~\cite{WahCUB_200_2011,yang2022banmo,wu2023dove,sinha2023common}, and more diverse datasets~\cite{XianAwA22019,yang2022apt,Ng_2022_CVPR,xu2023animal3d} often contain many noisy images unsuitable for training off the shelf.
To train our pan-category model for a wide range of quadruped animal species, we aggregate these existing datasets after substantial filtering, and additionally source more images from the Internet to create a large-scale object-centric image dataset spanning over $100$ quadruped species, as detailed in \cref{sec:dataset}.

\section{Method}%
\label{s:method}

Our goal is to learn a deformable model of a large variety of different animals using only Internet images for supervision.
Formally, we learn a function $f : I \mapsto O$ that maps any image $I \in \mathbb{R}^{3\times H\times W}$ of an animal to a corresponding 3D reconstruction $O$, capturing the animal's shape, deformation and appearance.

3D reconstruction is greatly facilitated by using multi-view data~\cite{hartley04multiple}, but this is not available at scale, or at all, for most animals.
Instead, we wish to reconstruct animals from weak single-view supervision obtained from the Internet.
Compared to prior works~\cite{wu2023magicpony,yao2022lassie,yao2023hi,yao2023artic3d}, which focused on reconstructing a single animal type at a time, here we target a large number of animal species at once, which is significantly more difficult.
We show in the next section how solving this problem requires carefully exploiting the semantic similarities and geometric correspondences between different animals to regularize their 3D geometry.

\subsection{Semantic Bank of Skinned Models}%
\label{s:sbsm}

Given an image $I$, consider the problem of estimating the 3D shape $(V, F)$ of the animal contained in it, where $V \in \mathbb{R}^{K\times 3}$ is a list of vertices of a 3D mesh with face connectivity given by triplets $F \subset \{1,\dots,K\}^3$.
While recovering a 3D shape from a single image is ill-posed, as we train the model $f$ on a large dataset, we can ultimately observe animals from a variety of viewpoints.
However, different images show different animals with different 3D shapes.
Non-Rigid Structure-from-Motion~\cite{Bregler00nrsfm,torresani2003learning,tretschk2023state} shows that reconstruction is still possible, but only if one  makes the space of possible 3D shapes sufficiently \emph{tight} to remove the reconstruction ambiguity.
At the same time, the space must be sufficiently \emph{expressive} to capture all animals.

\begin{figure}[t]
\begin{center}
\includegraphics[trim={20px 5px 0px 0px}, clip, width=0.85\linewidth]{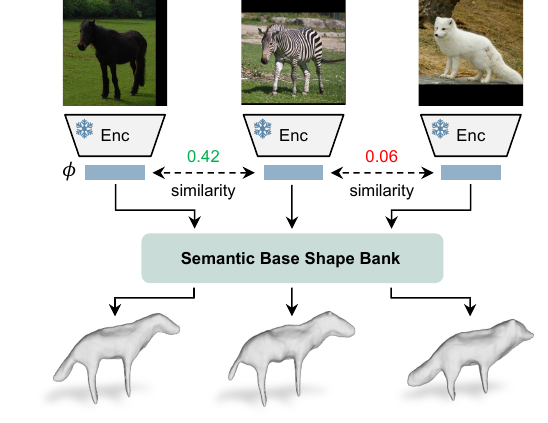}
\end{center}
\vspace{-0.25in}
\caption{\textbf{Queries from the Semantic Base Shape Bank.}
Without requiring any category labels, the Semantic Bank (Sec~\ref{s:sbsm}) automatically learns diverse base shapes for various animals and preserves the semantic similarities across different instances.
}
\vspace{-0.1in}
\label{fig:case-similarity}
\end{figure}

\paragraph{Skinned Models (SM).}

Following SMPL~\cite{loper2015smpl}, many works~\cite{wu2023dove, wu2023magicpony, yang2023rac, jakab2023farm3d} have adopted a Skinned Model (SM) to model the shape of deformable objects when learning from single-view image collections or videos.
An SM starts from a base shape $V_\text{base}$ of the object (\eg, human or animal) at `rest', applies as a \emph{small} deformation $V_\text{ins} = f_\text{ins}(V_\text{base}, \phi)$ to capture instance-specific details, and then applies a larger deformation via a skinning function $V = f_\text{pose}(V_\text{ins}, \phi)$, controlled by the articulation of the underlying skeleton.
We assume that deformations are predicted by neural networks that receive as input image features $\phi = f_{\phi}(I)$ extracted from a powerful self-supervised image encoder.

In our case, a single SM is insufficient to capture the very large shape variations between different animals, which include horses, dogs, antelopes, hedgehogs, \etc.
Na{\"\i}vely attempting to capture this diversity using the network $f_\text{ins}$ means that the resulting deformations \emph{cannot be small} any longer, which throws off the tightness of the model.

\paragraph{Semantic Bank of Skinned Models.}

In order to increase the expressiveness of the model while still avoiding overfitting individual images, we propose to exploit the fact that different animals often have similar 3D shapes as a result of evolution.
We can thus reduce the shape variation to a small number of shape bases $V_\text{base}$, and interpolate between them.

To do so, we introduce a \emph{Semantic Bank of Skinned Models} that automatically discovers a set of latent shape bases and learns to project each image into a linear combination of these bases.
Key to this method is to use pre-trained unsupervised image features~\cite{caron2021emerging,oquab2023dinov2} to automatically and implicitly identify similar animals.
This is realized by means of a small memory bank with $K$ learned key-value pairs $\{(\phi^\text{key}_k, \phi^\text{val}_k)\}_{k=1}^K$.
Specifically, given an image embedding $\phi$, we query the memory bank to obtain a latent shape embedding $\tilde{\phi}$ as a linear combination of the value tokens $\{\phi^\text{val}_k\}$ via a mechanism similar to attention~\cite{vaswani2017attention}:
\begin{equation}
    \tilde{\phi} = \sum_{k=1}^{K} w_k\, \phi^\text{val}_k, \; \text{where} \;
    w_k = \frac{\operatorname{cossim}(\phi, \phi^\text{key}_k)}
    {\sum_{j=1}^{K}\operatorname{cossim}(\phi, \phi^\text{key}_j)},
\end{equation}
and $\operatorname{cossim}$ denotes cosine similarity between two feature vectors.
This embedding $\tilde{\phi}$ is then used as a condition to the base shape predictor $(V_\text{base}, F) = f_\text{s}(\tilde{\phi})$, which produces semantically-adaptive base shapes without relying on any category labels or being bound to a hard categorization.

In practice, the image features $\phi$ are obtained from a well-trained feature extractor like DINO-ViT~\cite{caron2021emerging,oquab2023dinov2}.
Defining the weights based on the cosine similarities between the image features $\phi$ and a small number of bases $\{\phi^\text{key}_k\}$ captures the semantic similarities across different animal instances.
For instance, as illustrated in \cref{fig:case-similarity}, the cosine similarity between the image features of a zebra and a horse is $0.42$, whereas the similarity between a zebra and an arctic fox is only $0.06$.
Ablations in \cref{fig:ablation} further verify the importance of this Semantic Bank, without which the model easily overfits each training image and fails to reconstruct plausible 3D shapes.

\paragraph{Implementation Details.}

The base shape is predicted using a hybrid SDF-mesh representation~\cite{shen2021dmtet,wu2023magicpony} parameterized by a coordinate MLP,
with a conditioning vector $\tilde{\phi}$ injected via layer weight modulation~\cite{karras2020analyzing,karras2021alias}.
Since extracting meshes from SDFs using DMTet~\cite{shen2021dmtet} is memory and compute intensive, in practice, we only compute it once for each iteration, by assuming the batched images contain the same animal species, and simply averaging out the embeddings $\tilde{\phi}$.
The instance-specific deformation is predicted using another coordinate MLP that outputs the displacement $\Delta V_{\text{ins},i} = f_{\Delta V}(V_{\text{base},i}, \phi)$ for each vertex $V_{\text{base},i}$ of the base mesh conditioned on the image feature $\phi$, resulting in the deformed shape $V_\text{ins} = \Delta V_\text{ins} + V_\text{base}$.
We enforce a bilateral symmetry on both the base shape and the instance deformation by mirroring the query locations for the MLPs.
Given the instance mesh $V_\text{ins}$, we initialize a quadrupedal skeleton using a simple heuristic~\cite{wu2023magicpony}, and predict the rigid pose $\xi_1 \in SE(3)$ and bone rotations $\xi_b \in SO(3), b=2, \ldots, B$ using a pose network.
These posing parameters are then applied to the instance mesh via a linear blend skinning equation~\cite{loper2015smpl}.
Refer to the sup.~mat.~for more details.

\paragraph{Appearance.}

Assuming a Lambertian illumination model, we model the appearance of the object using an albedo field $a(\bx) = f_\text{a}(\bx, \phi) \in [0,1]^3$ and a dominant directional light.
The final shaded color of each pixel is computed as
$
\hat{I}(\bu) = \left( k_a + k_d \cdot \max \{0, \langle \bl, \bn \rangle \} \right) \cdot a(\bx)
$,
where $\bn$ is the normal direction of the \emph{posed} mesh at pixel $\bu$, and $k_a, k_d \in [0,1]$ and $\bl \in \mathbb{S}^2$ are respectively the ambient intensity, diffuse intensity and dominant light direction predicted by the lighting network $(k_a, k_d, \bl) = f_\text{l}(\phi)$.

\subsection{Learning Formulation}

The entire pipeline is trained in an unsupervised fashion, using only self-supervised image features~\cite{caron2021emerging, oquab2023dinov2} and object masks obtained from off-the-shelf segmenters~\cite{kirillov2023segment, kirillov2020pointrend}.

\paragraph{Reconstruction Losses.}

Given the final predicted posed shape $V$ and appearance of the object, we use a differentiable renderer $\cR$ to obtain an RGB image $\hat{I}$ as well as a mask image $\hat{M}$, which are compared to the input image $I$ and the pseudo-ground-truth object mask $M$:
\begin{align}
    \cL_\text{m} &= \|\hat{M} - M\|_2^2 + \lambda_\text{dt} \|\hat{M} \odot \texttt{dt}(M)\|_1, \\
    \cL_\text{im} &= \|\tilde{M} \odot (\hat{I} - I)\|_1,
\end{align}
where $\texttt{dt}(\cdot)$ is distance transform for more effective gradients~\cite{kanazawa18cmr,wu2021rendering}, $\odot$ denotes the Hadamard product, $\lambda_\text{dt}$ specifies the balancing weight, and $\tilde{M} = \hat{M} \odot M$ is the intersection of the predicted and ground-truth masks.

\paragraph{Correspondences from Self-Supervised Features.}

Self-supervised feature extractors are notoriously good at establishing semantic correspondences between objects, which can be distilled to facilitate 3D reconstruction~\cite{wu2023magicpony}.
To do so, we extract a patch-based feature map $\Phi \in \mathbb{R}^{D \times H \times W}$ from each training image.
These raw feature maps can be noisy and may preserve image-specific information irrelevant to other images.
To distill more effective semantic correspondences across different images, we perform a Principal Component Analysis (PCA) across all feature maps~\cite{wu2023magicpony}, reducing the dimension to $D' = 16$.
We then task the model to also learn a feature field in the canonical frame $\psi(\bx, \tilde{\phi}) \in \mathbb{R}^{D'}$ that is rendered into a feature image $\hat{\Phi}$ given predicted posed shape using the same renderer $\cR$.
Training then encourages the rendered feature images $\hat{\Phi}$ to match the pre-extracted PCA features $\Phi'$:
$
\cL_\text{feat} = \|\tilde{M} \odot (\hat{\Phi} - \Phi')\|_2^2.
$
Note that although the space of the PCA features $\Phi'$ is shared across different animal instances, the feature field $\psi$ still receives the latent embedding $\tilde{\phi}$ as a condition.
This is because different animals vary in shape, resulting in different feature fields.

\paragraph{Mask Discriminator.}

In practice, despite exploiting these semantic correspondences, we still find that the viewpoint prediction may easily collapse to only frontal viewpoints, due to the heavy photographer bias in Internet photos.
This can lead to overly elongated shapes as shown in \cref{fig:ablation}, and further deteriorates the viewpoint predictions.
To mitigate this, we further encourage the shape to look realistic from arbitrary viewpoints.
Specifically, we introduce a mask discriminator $D$ that encourages the mask images $\hat{M}_\text{rv}$ rendered from a random viewpoint to stay within the distribution of the ground-truth masks $\cM$.
The discriminator also receives the base embedding $\tilde{\phi}$ (with gradients detached) as a condition to make this adversarial guidance tailored to specific types of animals and thus more effective.
Formally, this is achieved via an adversarial loss~\cite{goodfellow2014generative}:
\begin{multline}
\cL_\text{adv} = \mathbb{E}_{M \sim \cM} [ \log D(M; \tilde{\phi}) ] \\ +
\mathbb{E}_{\hat{M}_\text{rv} \sim \cM_\text{rv}} [ \log (1 - D(\hat{M}_\text{rv}; \tilde{\phi})) ].
\label{eq:adv}
\end{multline}

Note that we do not use a discriminator on the rendered RGB images, as the predicted texture is often much less realistic when compared to real images, which gives the discriminator a trivial task.
Moreover, the distribution of mask images is less susceptible to viewpoint bias than RGB images, and hence we can simply sample random viewpoints uniformly, without requiring a precise viewpoint distribution of the training images.

\paragraph{Overall Loss.}

We further enforce the Eikonal constraint $\cR_\text{Eik}$ on the SDF network as well as the viewpoint hypothesis loss $\cL_\text{hyp}$ and the magnitude regularizers $\cR_\text{def}$ on vertex deformations and $\cR_\text{art}$ on articulation parameters $\xi$.
See the supplementary materials for details.

The final training objective $\cL$ is thus
\begin{equation}
    \cL = \cL_\text{rec} +
    \lambda_\text{hyp} \cL_\text{hyp} +
    \lambda_\text{adv} \cL_\text{adv} + \cR,
\end{equation}
where $
\cL_\text{rec} = \lambda_\text{m} \cL_\text{m} +
\lambda_\text{im} \cL_\text{im} +
\lambda_\text{feat} \cL_\text{feat}
$
summarizes the three reconstruction losses,
$
\cR = \lambda_\text{Eik} \cR_\text{Eik} +
\lambda_\text{art} \cR_\text{art} +
\lambda_\text{def} \cR_\text{def}
$
summarizes the regularizers,
and $\lambda$'s balance the contribution of each term.

\paragraph{Training Schedule.}

We design a robust training schedule that comprises three stages.
First, we train the base shapes and the viewpoint network without articulation or deformation.
This significantly improves the stability of the training and allows the model to roughly register the rigid pose of all instances and learn the coarse base shapes.

As the viewpoint prediction stabilizes after $20$k iterations, in the second stage, we instantiate the bones and enable the articulation, allowing the shapes to gradually grow legs and fit the articulated pose in each image.
Meanwhile, we also turn on the mask discriminator to prevent viewpoint collapse and shape elongation.
In the final stage, we optimize the instance shape deformation field to allow the model to capture the fine-grained geometric details of individual instances, with the discriminator disabled, as it may corrupt the shape if overused.

\section{Dataset Collection}%
\label{sec:dataset}

In order to train this pan-category model for all types of quadruped animals, we create a new animal image dataset, dubbed the \textbf{Fauna Dataset}, spanning $128$ quadruped species from dogs, antelopes to minks and platypuses, with a total of $78,\!168$ images.
We first aggregate the training sets of existing animal image datasets, including Animals-with-Attributes~\cite{XianAwA22019}, APT-36K~\cite{yang2022apt}, Animal3D~\cite{xu2023animal3d} and DOVE~\cite{wu2023dove}.
Many of these images are blurry or contain heavy occlusions, which will impact the stability of the training.
We thus filter the images using automatic scripts first, followed by manual inspection.
This results in $8,\!378$ images covering approximately $70$ animal species.
To further increase the size as well as the diversity of the dataset, we additionally collect $69,\!790$ images from the Internet, including $63,\!115$ video frames and $2,\!358$ images for $7$ common animals (bear, cow, elephant, giraffe, horse, sheep, zebra) as well as $4,\!317$ images for another $51$ less common species.
We use off-the-shelf segmentation models~\cite{kirillov2020pointrend,kirillov2023segment} to detect and segment the instances in the images.
Out of the $121$ few-shot categories, we hold out $5$ as novel categories unused at training.
For validation, we randomly select $5$ images in each of the rest $116$ few-shot categories, and $2,\!462$ images for the $7$ common species.
To reduce the viewpoint bias in the few-shot categories, we manually identify a few (1--10) backward-facing instances in the training set and duplicate them to match the size of the rest.

\begin{figure*}
    \centering
    \includegraphics[width=0.99\linewidth]{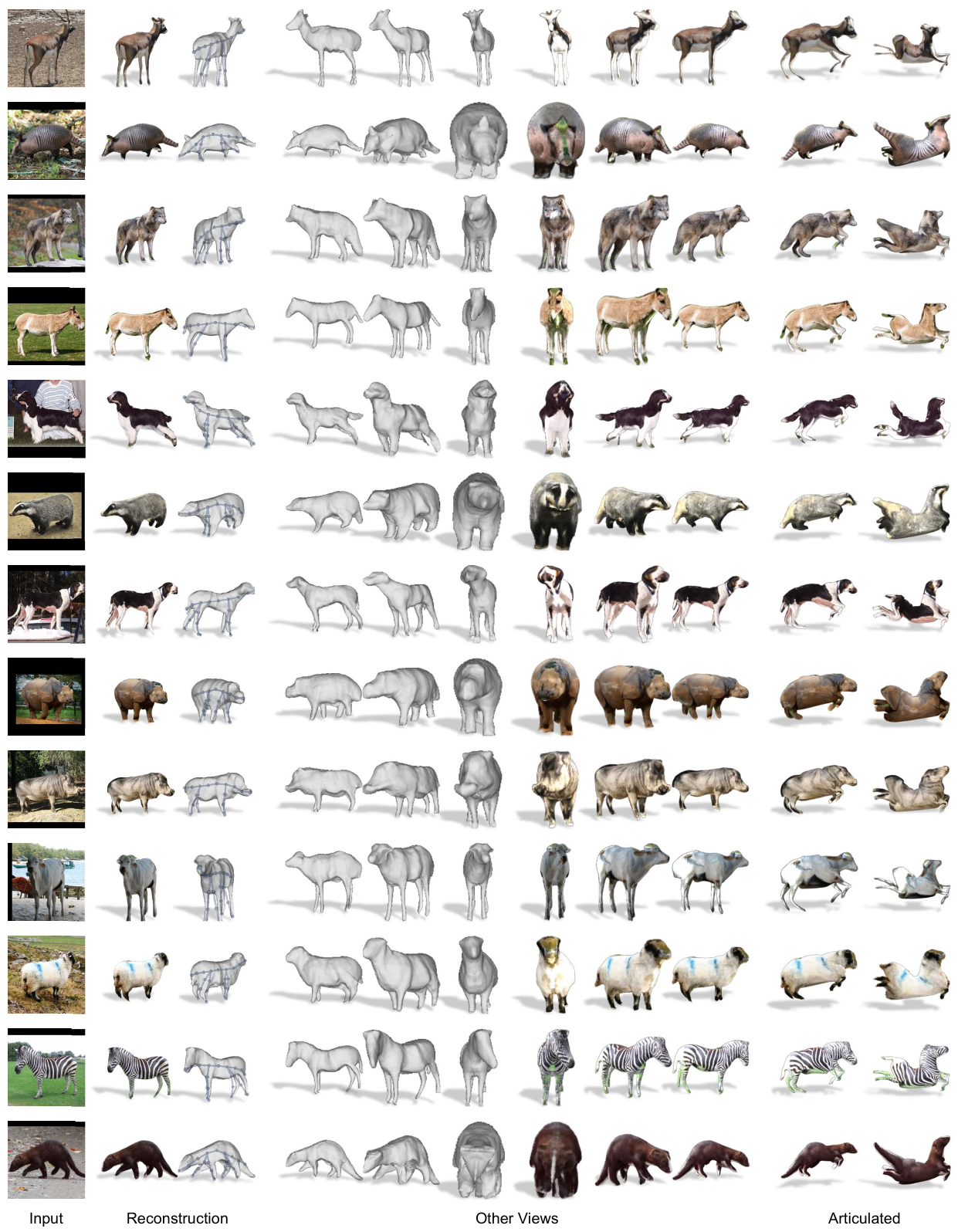}
    \caption{\textbf{Single Image 3D Reconstruction.}
    Given a single image of any quadruped animal at test time, our model reconstructs an articulated and textured 3D mesh in a feed-forward manner without requiring category labels, which can be readily animated.}
    \label{fig:exp-main-results}
\end{figure*}

\section{Experiments}%
\label{s:experiments}

\begin{figure*}[t]
    \centering
    \includegraphics[trim={0px 10px 0px 0}, width=\linewidth]{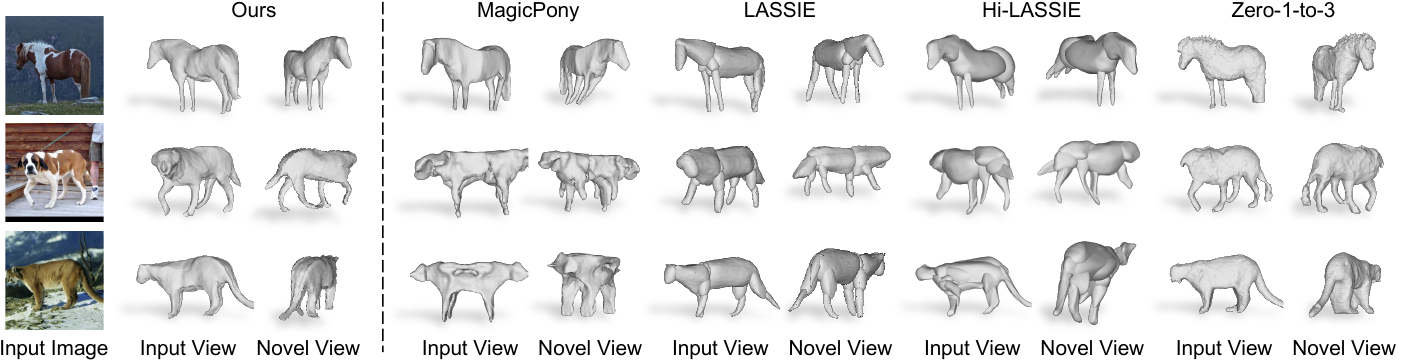}
    \vspace{-1em}
    \caption{\textbf{Qualitative Comparisons} against MagicPony~\cite{wu2023magicpony}, LASSIE~\cite{yao2022lassie}, Hi-LASSIE~\cite{yao2023hi} and Zero-1-to-3~\cite{liu2023zero}.
    Compared to all baselines, our method predicts more stable poses and higher-fidelity reconstructions.
    Note that our method is learning-based and predicts 3D meshes in a feed-forward fashion (as opposed to~\cite{yao2022lassie, yao2023hi} that optimize on test images),
    which is orders of magnitude faster.}
    \vspace{-0.1in}
    \label{fig:exp-compare-results}
\end{figure*}

\begin{figure}[t]
\begin{center}
\includegraphics[trim={0px 20px 0px 0}, width=\linewidth]{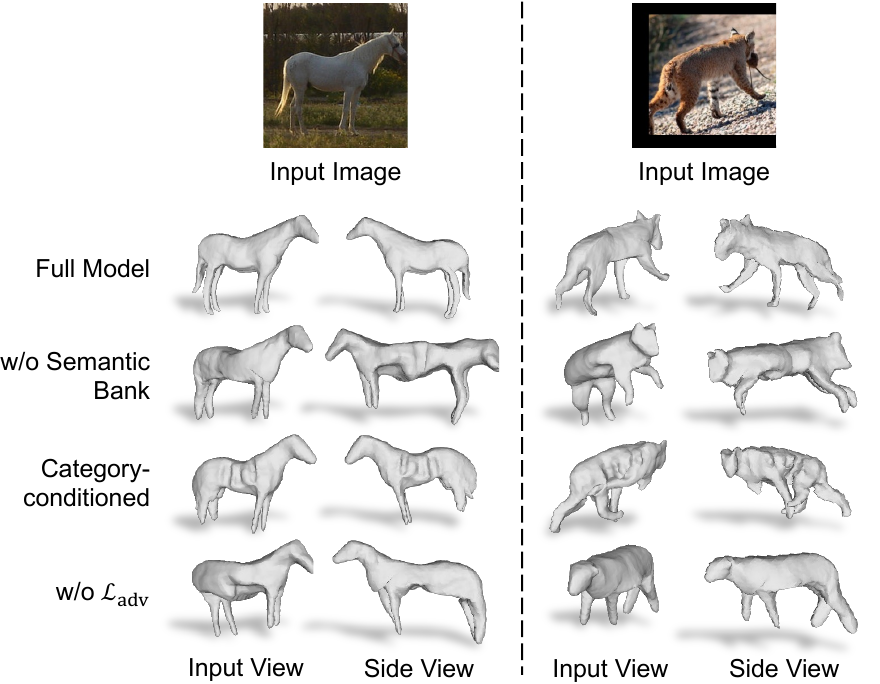}
\end{center}
    \vspace{-0.5em}
    \caption{\textbf{Ablation Studies.}
    Both the Semantic Bank and the mask discriminator improve the results as discussed in \cref{s:ablation}.}
    \vspace{-0.2in}
\label{fig:ablation}
\end{figure}

\subsection{Technical Details}

We base our architecture on MagicPony~\cite{wu2023magicpony}, adding the new SBSM and mask discriminator.
For the Semantic Bank, we use $K=60$ key-value pairs.
The dimension of keys is 384 (same as DINO-ViT) and the dimension of values is 128.
As the texture network tends to struggle to predict detailed appearance in one go, partially due to limited capacity, for all the visualizations, we follow~\cite{wu2023magicpony} and fine-tune (only) the texture network for $50$ iterations, which takes $<10$ seconds. Refer to the sup.~mat. for further details.

\subsection{Qualitative Results}

After training, \method takes in a \text{single} test image of any quadruped animal and produces an articulated and textured 3D mesh in a feed-forward manner, as visualized in \cref{fig:exp-main-results}.
The model can reconstruct very different animals, such as antelopes, armadillos, and fishers, without requiring any category labels.
All the input images in \cref{fig:exp-main-results} have not been seen during training.
In particular, the model also performs well on held-out categories, \eg the wolf in the third row.

\subsection{Comparisons with Prior Work}

\paragraph{Baselines.}

To the best of our knowledge, ours is the first deformable model designed to handle $100$+ quadruped species, learned purely from 2D Internet data.
We carry out quantitative and qualitative comparisons to methods that are at least in principle applicable to this setting.
The baseline is MagicPony~\cite{wu2023magicpony}, which however is \emph{category-specific} (they first train on horses, and fine-tune on giraffes, cows and zebras).
We also compare with two popular deformable models that can work in the wild, namely UMR~\cite{li2020self} and A-CSM~\cite{kulkarni2020articulation}.
However, they require weakly-supervised part segmentations and shape templates, respectively.
Other works, such as LASSIE~\cite{yao2022lassie} and its follow-ups~\cite{yao2023hi, yao2023artic3d}, optimize a deformable model on a small set of about 20 images covering a single animal category at a time.
More recently, image-to-3D methods based on distilling 2D diffusion models and/or large 3D datasets~\cite{liu2023zero} have also demonstrated plausible 3D reconstructions of animals from a single image.
In contrast, our model predicts an \emph{articulated} mesh from a single image within seconds.
Although it is difficult to establish a fair numerical comparison given these different settings, in \cref{sec:qual_comparison}, we provide a side-by-side qualitative comparison against baselines~\cite{yao2022lassie, yao2023hi, liu2023zero}.
We use the publicly released code~\cite{wu2023magicpony, yao2022lassie, yao2023hi, liu2023zero} and report numbers~\cite{li2020self, kulkarni2020articulation} included in MagicPony~\cite{wu2023magicpony}.

\begin{table}[t]
    \small
	\setlength{\tabcolsep}{0.06cm} %
    \renewcommand{\arraystretch}{1.1}
	\centering
	\begin{tabular}{lcccc}
\toprule
                               & \multicolumn{2}{c}{PASCAL}        & APT-36K         & Animal3D       \\ [-2pt]
                               \cmidrule(lr){2-3} \cmidrule(lr){4-4} \cmidrule(lr){5-5}
                               & {\footnotesize KT-PCK@0.1} & {\footnotesize PCK@0.1} & {\footnotesize PCK@0.1}        & {\footnotesize PCK@0.1}        \\ \midrule
UMR~\cite{li2020self}       & 0.284  & -           & -              & -              \\
A-CSM~\cite{kulkarni2020articulation}     & 0.329 & 0.687            & 0.649              & 0.822              \\
MagicPony~\cite{wu2023magicpony} & 0.429   & -          & 0.756          & 0.867          \\
Ours      & \textbf{0.539} & \textbf{0.782}    & \textbf{0.841} & \textbf{0.901} \\ \bottomrule
\end{tabular}
	\caption{
	\textbf{Quantitative Comparisons} on PASCAL VOC~\cite{everingham2015pascal}, APT-36K~\cite{yang2022apt} and Animal3D~\cite{xu2023animal3d}. When compared to baselines including the competitive MagicPony~\cite{wu2023magicpony}, our method demonstrates significantly improved performance on all datasets.}%
	\label{tab:all}
\vspace{-0.1in}
\end{table}

\paragraph{Quantitative Comparisons.}%
\label{s:quantitative}

We conduct quantitative evaluation across three different datasets, APT-36K~\cite{yang2022apt}, Animal3D~\cite{xu2023animal3d}, and PASCAL VOC~\cite{everingham2015pascal}, which contain images of various animals with 2D keypoint annotations.
Following MagicPony~\cite{wu2023magicpony}, we first evaluate on horses in PASCAL VOC~\cite{everingham2015pascal} using the widely used Keypoint Transfer metric~\cite{kanazawa18cmr,kulkarni2020articulation,li2020self}.
We use the same protocol as in A-CSM~\cite{kulkarni2020articulation} and randomly sample 20k source-target image pairs.
For each source image, we project the visible vertices of the predicted mesh onto the image and map each annotated 2D keypoint to its nearest vertex.
We then project that vertex to the target image and check if it lies within a small distance (10\% of image size) to the corresponding keypoint in the target image.
We summarize the results using the Percentage of Correct Keypoints~(KT-PCK@0.1) in \cref{tab:all}.

In \cref{tab:all}, we follow CMR~\cite{kanazawa18cmr} to evaluate the three datasets on more species, optimizing a linear mapping from mesh vertices to desired keypoints for each category,
and reporting PCK@0.1 between the predicted and annotated 2D keypoints.
Our model demonstrates significant improvement over existing methods on all datasets.
A performance breakdown for each category is provided in the sup.~mat.

\paragraph{Qualitative Comparisons.}%
\label{sec:qual_comparison}

\Cref{fig:exp-compare-results} compares \method qualitatively to several recent works~\cite{wu2023magicpony, yao2022lassie, yao2023hi, liu2023zero}.
To establish a fair comparison with MagicPony~\cite{wu2023magicpony}, for categories demonstrated in their paper (\eg horse), we simply run inference using the released model.
For each of the other categories, we use their public code to train a per-category model on our dataset from scratch (which contains less than 100 images for some rare categories).
For LASSIE~\cite{yao2022lassie} and Hi-LASSIE~\cite{yao2023hi}, which optimize over a small set of images, we train their models on the \emph{test} image together with additional $29$ images randomly selected from the training set of that category.
Hi-LASSIE~\cite{yao2023hi} is further fine-tuned on the test image after training.
To compare with Zero-1-to-3~\cite{liu2023zero}, we use the implementation in threestudio~\cite{threestudio2023} to first distill a NeRF~\cite{mildenhall2020nerf} using Score Distillation Sampling~\cite{poole2022dreamfusion} given the masked test image, and then extract a 3D mesh for fair comparison.
Note that our model predicts 3D meshes within seconds, whereas the optimization takes at least 10--20 mins for the other methods~\cite{yao2022lassie, yao2023hi, liu2023zero}.

As shown in \cref{fig:exp-compare-results}, MagicPony is sensitive to the size of the training set.
When trained on rare categories with fewer ($<100$) images, such as the puma in \cref{fig:exp-compare-results}, it fails to learn meaningful shapes and produces severe artifacts.
Despite optimizing on the test images, LASSIE and Hi-LASSIE produce coarser reconstructions, partially due to the part-based representation that struggles in capturing the detailed geometry and articulation, as well as unstable viewpoint prediction.
Zero-1-to-3, on the other hand, often fails to correctly reconstruct the legs, and does not explicitly model the articulated pose.
On the contrary, our method predicts accurate viewpoint and reconstructs fine-grained articulated shapes for all different animals, with only one \emph{single} model.

\subsection{Ablation Study}%
\label{s:ablation}

In \cref{fig:ablation}, we present ablation results on three key design choices in our pipeline: SBSM, category-agnostic training, and mask discriminator.
If we remove the SBSM and directly condition the base shape network on each individual image embedding $\phi$, the model tends to overfit each training views without learning meaningful canonical 3D shapes and pose.
Alternatively, we can simply condition the base shape on an explicit (learned) category-specific embedding and train the model in a category-conditioned manner.
This also leads to sub-optimal reconstructions, in particular on rare categories with few training images.
Lastly, training without the mask discriminator results in biased viewpoint prediction (towards frontal) and produces elongated shapes.
\section{Conclusions}%
\label{s:conclusion}

We have presented \method, a deformable model for 100 animal categories learned using only Internet images.
\method can reconstruct any quadruped image by instantiating in seconds a posed version of the deformable model to match the input image.
Despite capable of modeling diverse animals, the current model is still limited to quadruped species that share a same skeletal structure.
Furthermore, the training images still need to be lightly curated.
Nevertheless, \method still presents a significant leap compared to prior works and moves us closer to models that will be able to understand and reconstruct all animals in nature.

\paragraph{Acknowledgments.}

We thank Cristobal Eyzaguirre, Kyle Sargent, and Yunhao Ge for their insightful discussions and Chen Geng for proofreading. The work is in part supported by the Stanford Institute for Human-Centered AI (HAI), NSF RI \#2211258, \#2338203, ONR MURI N00014-22-1-2740, ONR YIP N00014-24-1-2117, the Samsung Global Research Outreach (GRO) program, Amazon, Google, and EPSRC VisualAI EP/T028572/1.

{
\small
\bibliographystyle{ieeenat_fullname}
\bibliography{ref,vedaldi_general,vedaldi_specific}

\begin{thebibliography}{81}
\providecommand{\natexlab}[1]{#1}
\providecommand{\url}[1]{\texttt{#1}}
\expandafter\ifx\csname urlstyle\endcsname\relax
  \providecommand{\doi}[1]{doi: #1}\else
  \providecommand{\doi}{doi: \begingroup \urlstyle{rm}\Url}\fi

\bibitem[Alwala et~al.(2022)Alwala, Gupta, and Tulsiani]{alwala2022pre}
Kalyan~Vasudev Alwala, Abhinav Gupta, and Shubham Tulsiani.
\newblock Pre-train, self-train, distill: A simple recipe for supersizing 3d
  reconstruction.
\newblock In \emph{CVPR}, 2022.

\bibitem[Ayg{\"u}n and Mac~Aodha(2024)]{aygun2023saor}
Mehmet Ayg{\"u}n and Oisin Mac~Aodha.
\newblock Saor: Single-view articulated object reconstruction.
\newblock In \emph{CVPR}, 2024.

\bibitem[Bogo et~al.(2016)Bogo, Kanazawa, Lassner, Gehler, Romero, and
  Black]{Bogo16}
Federica Bogo, Angjoo Kanazawa, Christoph Lassner, Peter Gehler, Javier Romero,
  and Michael~J. Black.
\newblock Keep it {SMPL}: Automatic estimation of {3D} human pose and shape
  from a single image.
\newblock In \emph{ECCV}, 2016.

\bibitem[Bregler et~al.(2000)Bregler, Hertzmann, and Biermann]{Bregler00nrsfm}
Christoph Bregler, Aaron Hertzmann, and Henning Biermann.
\newblock Recovering non-rigid 3d shape from image streams.
\newblock In \emph{CVPR}, 2000.

\bibitem[Caron et~al.(2021)Caron, Touvron, Misra, J{\'e}gou, Mairal,
  Bojanowski, and Joulin]{caron2021emerging}
Mathilde Caron, Hugo Touvron, Ishan Misra, Herv{\'e} J{\'e}gou, Julien Mairal,
  Piotr Bojanowski, and Armand Joulin.
\newblock Emerging properties in self-supervised vision transformers.
\newblock In \emph{ICCV}, 2021.

\bibitem[Cashman and Fitzgibbon(2012)]{Cashman12dolphin}
Thomas~J. Cashman and Andrew~W. Fitzgibbon.
\newblock What shape are dolphins? building 3d morphable models from 2d images.
\newblock \emph{IEEE TPAMI}, 2012.

\bibitem[Chan et~al.(2021)Chan, Monteiro, Kellnhofer, Wu, and
  Wetzstein]{chan2021piGAN}
Eric Chan, Marco Monteiro, Petr Kellnhofer, Jiajun Wu, and Gordon Wetzstein.
\newblock {pi-GAN}: Periodic implicit generative adversarial networks for
  3d-aware image synthesis.
\newblock In \emph{CVPR}, 2021.

\bibitem[Chan et~al.(2022)Chan, Lin, Chan, Nagano, Pan, {De Mello}, Gallo,
  Guibas, Tremblay, Khamis, Karras, and Wetzstein]{Chan2022}
Eric~R. Chan, Connor~Z. Lin, Matthew~A. Chan, Koki Nagano, Boxiao Pan, Shalini
  {De Mello}, Orazio Gallo, Leonidas Guibas, Jonathan Tremblay, Sameh Khamis,
  Tero Karras, and Gordon Wetzstein.
\newblock Efficient geometry-aware {3D} generative adversarial networks.
\newblock In \emph{CVPR}, 2022.

\bibitem[Deng et~al.(2023)Deng, Jiang, Qi, Yan, Zhou, Guibas, and
  Anguelov]{deng2022nerdi}
Congyue Deng, Chiyu~"Max'' Jiang, Charles~R. Qi, Xinchen Yan, Yin Zhou,
  Leonidas Guibas, and Dragomir Anguelov.
\newblock Nerdi: Single-view nerf synthesis with language-guided diffusion as
  general image priors.
\newblock In \emph{CVPR}, 2023.

\bibitem[Everingham et~al.(2015)Everingham, Eslami, Van~Gool, Williams, Winn,
  and Zisserman]{everingham2015pascal}
Mark Everingham, SM~Ali Eslami, Luc Van~Gool, Christopher~KI Williams, John
  Winn, and Andrew Zisserman.
\newblock The pascal visual object classes challenge: A retrospective.
\newblock \emph{IJCV}, 2015.

\bibitem[Felzenszwalb and Huttenlocher(2000)]{felzenszwalb2000efficient}
Pedro~F Felzenszwalb and Daniel~P Huttenlocher.
\newblock Efficient matching of pictorial structures.
\newblock In \emph{CVPR}, 2000.

\bibitem[Fischler and Elschlager(1973)]{fischler73the-representation}
Martin~A. Fischler and Robert~A. Elschlager.
\newblock The representation and matching of pictorial structures.
\newblock \emph{{IEEE} Trans. on Computers}, 1973.

\bibitem[Goel et~al.(2020)Goel, Kanazawa, and Malik]{Goel2020ucmr}
Shubham Goel, Angjoo Kanazawa, and Jitendra Malik.
\newblock Shape and viewpoints without keypoints.
\newblock In \emph{ECCV}, 2020.

\bibitem[Goel et~al.(2023)Goel, Pavlakos, Rajasegaran, Kanazawa, and
  Malik]{goel2023humans}
Shubham Goel, Georgios Pavlakos, Jathushan Rajasegaran, Angjoo Kanazawa, and
  Jitendra Malik.
\newblock Humans in 4d: Reconstructing and tracking humans with transformers.
\newblock In \emph{ICCV}, 2023.

\bibitem[Goodfellow et~al.(2014)Goodfellow, Pouget-Abadie, Mirza, Xu,
  Warde-Farley, Ozair, Courville, and Bengio]{goodfellow2014generative}
Ian Goodfellow, Jean Pouget-Abadie, Mehdi Mirza, Bing Xu, David Warde-Farley,
  Sherjil Ozair, Aaron Courville, and Yoshua Bengio.
\newblock Generative adversarial nets.
\newblock \emph{NeurIPS}, 2014.

\bibitem[Guo et~al.(2023)Guo, Liu, Shao, Laforte, Voleti, Luo, Chen, Zou, Wang,
  Cao, and Zhang]{threestudio2023}
Yuan-Chen Guo, Ying-Tian Liu, Ruizhi Shao, Christian Laforte, Vikram Voleti,
  Guan Luo, Chia-Hao Chen, Zi-Xin Zou, Chen Wang, Yan-Pei Cao, and Song-Hai
  Zhang.
\newblock threestudio: A unified framework for 3d content generation.
\newblock \url{https://github.com/threestudio-project/threestudio}, 2023.

\bibitem[Hartley and Zisserman(2004)]{hartley04multiple}
Richard Hartley and Andrew Zisserman.
\newblock \emph{Multiple View Geometry in Computer Vision}.
\newblock Cambridge University Press, ISBN: 0521540518, second edition, 2004.

\bibitem[Ho et~al.(2020)Ho, Jain, and Abbeel]{ho2020denoising}
Jonathan Ho, Ajay Jain, and Pieter Abbeel.
\newblock Denoising diffusion probabilistic models.
\newblock \emph{NeurIPS}, 2020.

\bibitem[Huang et~al.(2023)Huang, Jampani, Thai, Li, Stojanov, and
  Rehg]{huang2023shapeclipper}
Zixuan Huang, Varun Jampani, Anh Thai, Yuanzhen Li, Stefan Stojanov, and
  James~M Rehg.
\newblock Shapeclipper: Scalable 3d shape learning from single-view images via
  geometric and clip-based consistency.
\newblock In \emph{CVPR}, 2023.

\bibitem[Jakab et~al.(2024)Jakab, Li, Wu, Rupprecht, and
  Vedaldi]{jakab2023farm3d}
Tomas Jakab, Ruining Li, Shangzhe Wu, Christian Rupprecht, and Andrea Vedaldi.
\newblock Farm3d: Learning articulated 3d animals by distilling 2d diffusion.
\newblock In \emph{3DV}, 2024.

\bibitem[Joo et~al.(2019)Joo, Simon, Li, Liu, Tan, Gui, Banerjee, Godisart,
  Nabbe, Matthews, et~al.]{joo2019panoptic}
Hanbyul Joo, Tomas Simon, Xulong Li, Hao Liu, Lei Tan, Lin Gui, Sean Banerjee,
  Timothy Godisart, Bart Nabbe, Iain Matthews, et~al.
\newblock Panoptic studio: A massively multiview system for social interaction
  capture.
\newblock \emph{IEEE TPAMI}, 2019.

\bibitem[Kanazawa et~al.(2018{\natexlab{a}})Kanazawa, Tulsiani, Efros, and
  Malik]{kanazawa18cmr}
Angjoo Kanazawa, Shubham Tulsiani, Alexei~A. Efros, and Jitendra Malik.
\newblock Learning category-specific mesh reconstruction from image
  collections.
\newblock In \emph{ECCV}, 2018{\natexlab{a}}.

\bibitem[Kanazawa et~al.(2018{\natexlab{b}})Kanazawa, Tulsiani, Efros, and
  Malik]{kanazawa2018learning}
Angjoo Kanazawa, Shubham Tulsiani, Alexei~A Efros, and Jitendra Malik.
\newblock Learning category-specific mesh reconstruction from image
  collections.
\newblock In \emph{ECCV}, 2018{\natexlab{b}}.

\bibitem[Karras et~al.(2020)Karras, Laine, Aittala, Hellsten, Lehtinen, and
  Aila]{karras2020analyzing}
Tero Karras, Samuli Laine, Miika Aittala, Janne Hellsten, Jaakko Lehtinen, and
  Timo Aila.
\newblock Analyzing and improving the image quality of stylegan.
\newblock In \emph{CVPR}, 2020.

\bibitem[Karras et~al.(2021)Karras, Aittala, Laine, H{\"a}rk{\"o}nen, Hellsten,
  Lehtinen, and Aila]{karras2021alias}
Tero Karras, Miika Aittala, Samuli Laine, Erik H{\"a}rk{\"o}nen, Janne
  Hellsten, Jaakko Lehtinen, and Timo Aila.
\newblock Alias-free generative adversarial networks.
\newblock \emph{NeurIPS}, 2021.

\bibitem[Kim et~al.(2023)Kim, Lee, Choi, Jeong, Sohn, and
  Shin]{kim2023collaborative}
Subin Kim, Kyungmin Lee, June~Suk Choi, Jongheon Jeong, Kihyuk Sohn, and Jinwoo
  Shin.
\newblock Collaborative score distillation for consistent visual synthesis.
\newblock \emph{arXiv preprint arXiv:2307.04787}, 2023.

\bibitem[Kirillov et~al.(2020)Kirillov, Wu, He, and
  Girshick]{kirillov2020pointrend}
Alexander Kirillov, Yuxin Wu, Kaiming He, and Ross Girshick.
\newblock Pointrend: Image segmentation as rendering.
\newblock In \emph{CVPR}, 2020.

\bibitem[Kirillov et~al.(2023)Kirillov, Mintun, Ravi, Mao, Rolland, Gustafson,
  Xiao, Whitehead, Berg, Lo, et~al.]{kirillov2023segment}
Alexander Kirillov, Eric Mintun, Nikhila Ravi, Hanzi Mao, Chloe Rolland, Laura
  Gustafson, Tete Xiao, Spencer Whitehead, Alexander~C Berg, Wan-Yen Lo, et~al.
\newblock Segment anything.
\newblock In \emph{ICCV}, 2023.

\bibitem[Kulkarni et~al.(2020)Kulkarni, Gupta, Fouhey, and
  Tulsiani]{kulkarni2020articulation}
Nilesh Kulkarni, Abhinav Gupta, David~F Fouhey, and Shubham Tulsiani.
\newblock Articulation-aware canonical surface mapping.
\newblock In \emph{CVPR}, 2020.

\bibitem[Li et~al.(2020)Li, Liu, Kim, De~Mello, Jampani, Yang, and
  Kautz]{li2020self}
Xueting Li, Sifei Liu, Kihwan Kim, Shalini De~Mello, Varun Jampani, Ming-Hsuan
  Yang, and Jan Kautz.
\newblock Self-supervised single-view 3d reconstruction via semantic
  consistency.
\newblock In \emph{ECCV}, 2020.

\bibitem[Liu et~al.(2023{\natexlab{a}})Liu, Xu, Jin, Chen, Xu, Su,
  et~al.]{liu2023one}
Minghua Liu, Chao Xu, Haian Jin, Linghao Chen, Zexiang Xu, Hao Su, et~al.
\newblock One-2-3-45: Any single image to 3d mesh in 45 seconds without
  per-shape optimization.
\newblock \emph{NeurIPS}, 2023{\natexlab{a}}.

\bibitem[Liu et~al.(2023{\natexlab{b}})Liu, Wu, Van~Hoorick, Tokmakov,
  Zakharov, and Vondrick]{liu2023zero}
Ruoshi Liu, Rundi Wu, Basile Van~Hoorick, Pavel Tokmakov, Sergey Zakharov, and
  Carl Vondrick.
\newblock Zero-1-to-3: Zero-shot one image to 3d object.
\newblock In \emph{ICCV}, 2023{\natexlab{b}}.

\bibitem[Liu et~al.(2023{\natexlab{c}})Liu, Lin, Zeng, Long, Liu, Komura, and
  Wang]{liu2023syncdreamer}
Yuan Liu, Cheng Lin, Zijiao Zeng, Xiaoxiao Long, Lingjie Liu, Taku Komura, and
  Wenping Wang.
\newblock {SyncDreamer}: Generating multiview-consistent images from a
  single-view image.
\newblock \emph{arXiv preprint arXiv:2309.03453}, 2023{\natexlab{c}}.

\bibitem[Long et~al.(2023)Long, Guo, Lin, Liu, Dou, Liu, Ma, Zhang, Habermann,
  Theobalt, et~al.]{long2023wonder3d}
Xiaoxiao Long, Yuan-Chen Guo, Cheng Lin, Yuan Liu, Zhiyang Dou, Lingjie Liu,
  Yuexin Ma, Song-Hai Zhang, Marc Habermann, Christian Theobalt, et~al.
\newblock Wonder3d: Single image to 3d using cross-domain diffusion.
\newblock \emph{arXiv preprint arXiv:2310.15008}, 2023.

\bibitem[Loper et~al.(2015)Loper, Mahmood, Romero, Pons-Moll, and
  Black]{loper2015smpl}
Matthew Loper, Naureen Mahmood, Javier Romero, Gerard Pons-Moll, and Michael~J
  Black.
\newblock {SMPL}: A skinned multi-person linear model.
\newblock \emph{ACM TOG}, 2015.

\bibitem[Melas-Kyriazi et~al.(2023)Melas-Kyriazi, Laina, Rupprecht, and
  Vedaldi]{melas2023realfusion}
Luke Melas-Kyriazi, Iro Laina, Christian Rupprecht, and Andrea Vedaldi.
\newblock Realfusion: 360deg reconstruction of any object from a single image.
\newblock In \emph{CVPR}, 2023.

\bibitem[Mildenhall et~al.(2020)Mildenhall, Srinivasan, Tancik, Barron,
  Ramamoorthi, and Ng]{mildenhall2020nerf}
Ben Mildenhall, Pratul~P. Srinivasan, Matthew Tancik, Jonathan~T. Barron, Ravi
  Ramamoorthi, and Ren Ng.
\newblock {NeRF}: Representing scenes as neural radiance fields for view
  synthesis.
\newblock In \emph{ECCV}, 2020.

\bibitem[Ng et~al.(2022)Ng, Ong, Zheng, Ni, Yeo, and Liu]{Ng_2022_CVPR}
Xun~Long Ng, Kian~Eng Ong, Qichen Zheng, Yun Ni, Si~Yong Yeo, and Jun Liu.
\newblock Animal kingdom: A large and diverse dataset for animal behavior
  understanding.
\newblock In \emph{CVPR}, 2022.

\bibitem[Nguyen-Phuoc et~al.(2019)Nguyen-Phuoc, Li, Theis, Richardt, and
  Yang]{nguyen2019hologan}
Thu Nguyen-Phuoc, Chuan Li, Lucas Theis, Christian Richardt, and Yong-Liang
  Yang.
\newblock {HoloGAN}: Unsupervised learning of 3d representations from natural
  images.
\newblock In \emph{ICCV}, 2019.

\bibitem[Niemeyer and Geiger(2021)]{Niemeyer2020GIRAFFE}
Michael Niemeyer and Andreas Geiger.
\newblock {GIRAFFE}: Representing scenes as compositional generative neural
  feature fields.
\newblock In \emph{CVPR}, 2021.

\bibitem[Oquab et~al.(2023)Oquab, Darcet, Moutakanni, Vo, Szafraniec, Khalidov,
  Fernandez, Haziza, Massa, El-Nouby, et~al.]{oquab2023dinov2}
Maxime Oquab, Timoth{\'e}e Darcet, Th{\'e}o Moutakanni, Huy Vo, Marc
  Szafraniec, Vasil Khalidov, Pierre Fernandez, Daniel Haziza, Francisco Massa,
  Alaaeldin El-Nouby, et~al.
\newblock Dinov2: Learning robust visual features without supervision.
\newblock \emph{arXiv preprint arXiv:2304.07193}, 2023.

\bibitem[Poole et~al.(2023)Poole, Jain, Barron, and
  Mildenhall]{poole2022dreamfusion}
Ben Poole, Ajay Jain, Jonathan~T Barron, and Ben Mildenhall.
\newblock Dreamfusion: Text-to-3d using 2d diffusion.
\newblock \emph{ICLR}, 2023.

\bibitem[Qian et~al.(2023)Qian, Mai, Hamdi, Ren, Siarohin, Li, Lee,
  Skorokhodov, Wonka, Tulyakov, et~al.]{qian2023magic123}
Guocheng Qian, Jinjie Mai, Abdullah Hamdi, Jian Ren, Aliaksandr Siarohin, Bing
  Li, Hsin-Ying Lee, Ivan Skorokhodov, Peter Wonka, Sergey Tulyakov, et~al.
\newblock Magic123: One image to high-quality 3d object generation using both
  2d and 3d diffusion priors.
\newblock \emph{arXiv preprint arXiv:2306.17843}, 2023.

\bibitem[R{\"u}egg et~al.(2022)R{\"u}egg, Zuffi, Schindler, and
  Black]{ruegg2022barc}
Nadine R{\"u}egg, Silvia Zuffi, Konrad Schindler, and Michael~J Black.
\newblock Barc: Learning to regress 3d dog shape from images by exploiting
  breed information.
\newblock In \emph{CVPR}, 2022.

\bibitem[R{\"u}egg et~al.(2023)R{\"u}egg, Tripathi, Schindler, Black, and
  Zuffi]{ruegg2023bite}
Nadine R{\"u}egg, Shashank Tripathi, Konrad Schindler, Michael~J Black, and
  Silvia Zuffi.
\newblock Bite: Beyond priors for improved three-d dog pose estimation.
\newblock In \emph{CVPR}, 2023.

\bibitem[Shen et~al.(2021)Shen, Gao, Yin, Liu, and Fidler]{shen2021dmtet}
Tianchang Shen, Jun Gao, Kangxue Yin, Ming-Yu Liu, and Sanja Fidler.
\newblock Deep marching tetrahedra: a hybrid representation for high-resolution
  3d shape synthesis.
\newblock \emph{NeurIPS}, 2021.

\bibitem[Shi et~al.(2023)Shi, Wang, Ye, Long, Li, and Yang]{shi2023mvdream}
Yichun Shi, Peng Wang, Jianglong Ye, Mai Long, Kejie Li, and Xiao Yang.
\newblock Mvdream: Multi-view diffusion for 3d generation.
\newblock \emph{arXiv preprint arXiv:2308.16512}, 2023.

\bibitem[Siddiqui et~al.(2022)Siddiqui, Thies, Ma, Shan, Nie{\ss}ner, and
  Dai]{siddiqui2022texturify}
Yawar Siddiqui, Justus Thies, Fangchang Ma, Qi Shan, Matthias Nie{\ss}ner, and
  Angela Dai.
\newblock Texturify: Generating textures on 3d shape surfaces.
\newblock In \emph{ECCV}, 2022.

\bibitem[Sinha et~al.(2023)Sinha, Shapovalov, Reizenstein, Rocco, Neverova,
  Vedaldi, and Novotny]{sinha2023common}
Samarth Sinha, Roman Shapovalov, Jeremy Reizenstein, Ignacio Rocco, Natalia
  Neverova, Andrea Vedaldi, and David Novotny.
\newblock Common pets in 3d: Dynamic new-view synthesis of real-life deformable
  categories.
\newblock In \emph{CVPR}, 2023.

\bibitem[Song et~al.(2021)Song, Sohl-Dickstein, Kingma, Kumar, Ermon, and
  Poole]{song2020score}
Yang Song, Jascha Sohl-Dickstein, Diederik~P Kingma, Abhishek Kumar, Stefano
  Ermon, and Ben Poole.
\newblock Score-based generative modeling through stochastic differential
  equations.
\newblock In \emph{ICLR}, 2021.

\bibitem[Sun et~al.(2023)Sun, Zhang, Shao, Wang, Liu, Xie, and
  Liu]{sun2023dreamcraft3d}
Jingxiang Sun, Bo Zhang, Ruizhi Shao, Lizhen Wang, Wen Liu, Zhenda Xie, and
  Yebin Liu.
\newblock Dreamcraft3d: Hierarchical 3d generation with bootstrapped diffusion
  prior.
\newblock \emph{arXiv preprint arXiv:2310.16818}, 2023.

\bibitem[Tang et~al.(2023)Tang, Ren, Zhou, Liu, and
  Zeng]{tang2023dreamgaussian}
Jiaxiang Tang, Jiawei Ren, Hang Zhou, Ziwei Liu, and Gang Zeng.
\newblock Dreamgaussian: Generative gaussian splatting for efficient 3d content
  creation.
\newblock \emph{arXiv preprint arXiv:2309.16653}, 2023.

\bibitem[Torresani et~al.(2004)Torresani, Hertzmann, and
  Bregler]{torresani2003learning}
Lorenzo Torresani, Aaron Hertzmann, and Christoph Bregler.
\newblock Learning non-rigid 3d shape from 2d motion.
\newblock \emph{NeurIPS}, 2004.

\bibitem[Tretschk et~al.(2023)Tretschk, Kairanda, BR, Dabral, Kortylewski,
  Egger, Habermann, Fua, Theobalt, and Golyanik]{tretschk2023state}
Edith Tretschk, Navami Kairanda, Mallikarjun BR, Rishabh Dabral, Adam
  Kortylewski, Bernhard Egger, Marc Habermann, Pascal Fua, Christian Theobalt,
  and Vladislav Golyanik.
\newblock State of the art in dense monocular non-rigid 3d reconstruction.
\newblock In \emph{Comput. Graph. Forum}, pages 485--520, 2023.

\bibitem[Tulsiani et~al.(2020)Tulsiani, Kulkarni, and Gupta]{tulsiani2020imr}
Shubham Tulsiani, Nilesh Kulkarni, and Abhinav Gupta.
\newblock Implicit mesh reconstruction from unannotated image collections.
\newblock \emph{arXiv preprint arXiv:2007.08504}, 2020.

\bibitem[Vaswani et~al.(2017)Vaswani, Shazeer, Parmar, Uszkoreit, Jones, Gomez,
  Kaiser, and Polosukhin]{vaswani2017attention}
Ashish Vaswani, Noam Shazeer, Niki Parmar, Jakob Uszkoreit, Llion Jones,
  Aidan~N Gomez, {\L}ukasz Kaiser, and Illia Polosukhin.
\newblock Attention is all you need.
\newblock \emph{NeurIPS}, 2017.

\bibitem[Wah et~al.(2011)Wah, Branson, Welinder, Perona, and
  Belongie]{WahCUB_200_2011}
Catherine Wah, Steve Branson, Peter Welinder, Pietro Perona, and Serge
  Belongie.
\newblock {The Caltech-UCSD Birds-200-2011 Dataset}.
\newblock Technical Report CNS-TR-2011-001, California Institute of Technology,
  2011.

\bibitem[Wang et~al.(2021)Wang, Kolotouros, Daniilidis, and
  Badger]{wang2021birds}
Yufu Wang, Nikos Kolotouros, Kostas Daniilidis, and Marc Badger.
\newblock Birds of a feather: Capturing avian shape models from images.
\newblock In \emph{CVPR}, 2021.

\bibitem[Weng et~al.(2023)Weng, Yang, Wang, Li, Zhang, Chen, and
  Zhang]{weng2023consistent123}
Haohan Weng, Tianyu Yang, Jianan Wang, Yu Li, Tong Zhang, CL Chen, and Lei
  Zhang.
\newblock Consistent123: Improve consistency for one image to 3d object
  synthesis.
\newblock \emph{arXiv preprint arXiv:2310.08092}, 2023.

\bibitem[Wu et~al.(2020)Wu, Rupprecht, and Vedaldi]{wu2020unsupervised}
Shangzhe Wu, Christian Rupprecht, and Andrea Vedaldi.
\newblock Unsupervised learning of probably symmetric deformable 3d objects
  from images in the wild.
\newblock In \emph{CVPR}, 2020.

\bibitem[Wu et~al.(2021)Wu, Makadia, Wu, Snavely, Tucker, and
  Kanazawa]{wu2021rendering}
Shangzhe Wu, Ameesh Makadia, Jiajun Wu, Noah Snavely, Richard Tucker, and
  Angjoo Kanazawa.
\newblock De-rendering the world's revolutionary artefacts.
\newblock In \emph{CVPR}, 2021.

\bibitem[Wu et~al.(2023{\natexlab{a}})Wu, Jakab, Rupprecht, and
  Vedaldi]{wu2023dove}
Shangzhe Wu, Tomas Jakab, Christian Rupprecht, and Andrea Vedaldi.
\newblock {DOVE}: Learning deformable 3d objects by watching videos.
\newblock \emph{IJCV}, 2023{\natexlab{a}}.

\bibitem[Wu et~al.(2023{\natexlab{b}})Wu, Li, Jakab, Rupprecht, and
  Vedaldi]{wu2023magicpony}
Shangzhe Wu, Ruining Li, Tomas Jakab, Christian Rupprecht, and Andrea Vedaldi.
\newblock Magicpony: Learning articulated 3d animals in the wild.
\newblock In \emph{CVPR}, 2023{\natexlab{b}}.

\bibitem[Wu et~al.(2023{\natexlab{c}})Wu, Li, Jakab, Rupprecht, and
  Vedaldi]{wu23magicpony}
Shangzhe Wu, Ruining Li, Tomas Jakab, Christian Rupprecht, and Andrea Vedaldi.
\newblock {MagicPony}: Learning articulated {3D} animals in the wild.
\newblock In \emph{Proceedings of the {IEEE} Conference on Computer Vision and
  Pattern Recognition ({CVPR})}, 2023{\natexlab{c}}.

\bibitem[Xian et~al.(2019)Xian, Lampert, Schiele, and Akata]{XianAwA22019}
Yongqin Xian, Christoph~H. Lampert, Bernt Schiele, and Zeynep Akata.
\newblock Zero-shot learning—a comprehensive evaluation of the good, the bad
  and the ugly.
\newblock \emph{IEEE TPAMI}, 2019.

\bibitem[Xu et~al.(2023{\natexlab{a}})Xu, Zhang, Peng, Ma, Jesslen, Ji, Hu,
  Zhang, Liu, Wang, et~al.]{xu2023animal3d}
Jiacong Xu, Yi Zhang, Jiawei Peng, Wufei Ma, Artur Jesslen, Pengliang Ji, Qixin
  Hu, Jiehua Zhang, Qihao Liu, Jiahao Wang, et~al.
\newblock Animal3d: A comprehensive dataset of 3d animal pose and shape.
\newblock In \emph{ICCV}, 2023{\natexlab{a}}.

\bibitem[Xu et~al.(2023{\natexlab{b}})Xu, Tan, Luan, Bi, Peng, Li, Shi,
  Sunkavalli, Gordon, Xu, and Kai]{dmv2023}
Yinghao Xu, Hao Tan, Fujun Luan, Sai Bi, Wang Peng, Jihao Li, Zifan Shi, Kaylan
  Sunkavalli, Wetzstein Gordon, Zexiang Xu, and Zhang Kai.
\newblock {DMV3D}: Denoising multi-view diffusion using 3d large reconstruction
  model.
\newblock \emph{arXiv preprint arXiv:2311.09217}, 2023{\natexlab{b}}.

\bibitem[Yang et~al.(2021{\natexlab{a}})Yang, Sun, Jampani, Vlasic, Cole,
  Chang, Ramanan, Freeman, and Liu]{yang21lasr}
Gengshan Yang, Deqing Sun, Varun Jampani, Daniel Vlasic, Forrester Cole, Huiwen
  Chang, Deva Ramanan, William~T. Freeman, and Ce Liu.
\newblock {LASR}: Learning articulated shape reconstruction from a monocular
  video.
\newblock In \emph{CVPR}, 2021{\natexlab{a}}.

\bibitem[Yang et~al.(2021{\natexlab{b}})Yang, Sun, Jampani, Vlasic, Cole, Liu,
  and Ramanan]{yang2021viser}
Gengshan Yang, Deqing Sun, Varun Jampani, Daniel Vlasic, Forrester Cole, Ce
  Liu, and Deva Ramanan.
\newblock {ViSER}: Video-specific surface embeddings for articulated 3d shape
  reconstruction.
\newblock In \emph{NeurIPS}, 2021{\natexlab{b}}.

\bibitem[Yang et~al.(2022{\natexlab{a}})Yang, Vo, Natalia, Ramanan, Andrea, and
  Hanbyul]{yang2022banmo}
Gengshan Yang, Minh Vo, Neverova Natalia, Deva Ramanan, Vedaldi Andrea, and Joo
  Hanbyul.
\newblock {BANMo}: Building animatable 3d neural models from many casual
  videos.
\newblock In \emph{CVPR}, 2022{\natexlab{a}}.

\bibitem[Yang et~al.(2023{\natexlab{a}})Yang, Wang, Reddy, and
  Ramanan]{yang2023rac}
Gengshan Yang, Chaoyang Wang, N.~Dinesh Reddy, and Deva Ramanan.
\newblock Reconstructing animatable categories from videos.
\newblock In \emph{CVPR}, 2023{\natexlab{a}}.

\bibitem[Yang et~al.(2023{\natexlab{b}})Yang, Cheng, Duan, Ji, and
  Li]{yang2023consistnet}
Jiayu Yang, Ziang Cheng, Yunfei Duan, Pan Ji, and Hongdong Li.
\newblock Consistnet: Enforcing 3d consistency for multi-view images diffusion.
\newblock \emph{arXiv preprint arXiv:2310.10343}, 2023{\natexlab{b}}.

\bibitem[Yang et~al.(2022{\natexlab{b}})Yang, Yang, Xu, Zhang, Lan, and
  Tao]{yang2022apt}
Yuxiang Yang, Junjie Yang, Yufei Xu, Jing Zhang, Long Lan, and Dacheng Tao.
\newblock Apt-36k: A large-scale benchmark for animal pose estimation and
  tracking.
\newblock \emph{NeurIPS}, 2022{\natexlab{b}}.

\bibitem[Yao et~al.(2022)Yao, Hung, Li, Rubinstein, Yang, and
  Jampani]{yao2022lassie}
Chun-Han Yao, Wei-Chih Hung, Yuanzhen Li, Michael Rubinstein, Ming-Hsuan Yang,
  and Varun Jampani.
\newblock Lassie: Learning articulated shapes from sparse image ensemble via 3d
  part discovery.
\newblock \emph{NeurIPS}, 2022.

\bibitem[Yao et~al.(2023{\natexlab{a}})Yao, Hung, Li, Rubinstein, Yang, and
  Jampani]{yao2023hi}
Chun-Han Yao, Wei-Chih Hung, Yuanzhen Li, Michael Rubinstein, Ming-Hsuan Yang,
  and Varun Jampani.
\newblock Hi-lassie: High-fidelity articulated shape and skeleton discovery
  from sparse image ensemble.
\newblock In \emph{CVPR}, 2023{\natexlab{a}}.

\bibitem[Yao et~al.(2023{\natexlab{b}})Yao, Raj, Hung, Li, Rubinstein, Yang,
  and Jampani]{yao2023artic3d}
Chun-Han Yao, Amit Raj, Wei-Chih Hung, Yuanzhen Li, Michael Rubinstein,
  Ming-Hsuan Yang, and Varun Jampani.
\newblock Artic3d: Learning robust articulated 3d shapes from noisy web image
  collections.
\newblock \emph{NeurIPS}, 2023{\natexlab{b}}.

\bibitem[Ye et~al.(2021)Ye, Tulsiani, and Gupta]{ye21shelf}
Yufei Ye, Shubham Tulsiani, and Abhinav Gupta.
\newblock Shelf-supervised mesh prediction in the wild.
\newblock In \emph{CVPR}, 2021.

\bibitem[Zhang et~al.(2023)Zhang, Wu, Snavely, and Wu]{zhang2023seeing}
Yunzhi Zhang, Shangzhe Wu, Noah Snavely, and Jiajun Wu.
\newblock Seeing a rose in five thousand ways.
\newblock In \emph{CVPR}, 2023.

\bibitem[Zhu et~al.(2017)Zhu, Park, Isola, and Efros]{zhu2017unpaired}
Jun-Yan Zhu, Taesung Park, Phillip Isola, and Alexei~A Efros.
\newblock Unpaired image-to-image translation using cycle-consistent
  adversarial networks.
\newblock In \emph{ICCV}, 2017.

\bibitem[Zuffi et~al.(2017)Zuffi, Kanazawa, Jacobs, and Black]{zuffi20173d}
Silvia Zuffi, Angjoo Kanazawa, David~W Jacobs, and Michael~J Black.
\newblock 3d menagerie: Modeling the 3d shape and pose of animals.
\newblock In \emph{CVPR}, 2017.

\bibitem[Zuffi et~al.(2018)Zuffi, Kanazawa, and Black]{zuffi2018lions}
Silvia Zuffi, Angjoo Kanazawa, and Michael~J Black.
\newblock Lions and tigers and bears: Capturing non-rigid, 3d, articulated
  shape from images.
\newblock In \emph{CVPR}, 2018.

\end{thebibliography}
}

\ifarxiv
\newpage

\appendix

\section{Additional Results}
We provide additional visualizations, including shape interpolation and generation, as well as additional comparisons in this supplementary material.
Please see {\footnotesize\url{https://kyleleey.github.io/3DFauna/}} for 3D animations.

\subsection{Shape Interpolation between Instances}
With the predictions of our model, we can easily interpolate between two reconstructions by interpolating the base embeddings $\tilde{\phi}$, instance deformations and the articulated poses $\xi$, as illustrated in \cref{fig:supp-interpolation}.
Here, we first obtain the predicted base shape embeddings $\tilde{\phi}$ for each of the three input images from the learned Semantic Bank.
We then linearly interpolate between these embeddings to produce smooth a transition from one base shape to another, as shown in the last row of \cref{fig:supp-interpolation}.
Furthermore, we can also linearly interpolate the predicted articulated the image features $\phi$ (which is used as a condition to the instance deformation field $f_{\Delta V}$) as well as the predicted articulation parameters $\xi$, to generate smooth interpolations of between posed shapes, shown in the middle row.
These results confirm that our learned shape space is continuous and smooth, and covers a wide range of animal shapes.

\subsection{Shape Generation from the Semantic Bank}
Moreover, we can also \emph{generate} new animal shapes by sampling from the learned Semantic Bank, as shown in \cref{fig:supp-sampling}.
First, we visualize the base shapes captured by each of the learned value tokens $\phi^\text{val}_k$ in the Semantic Bank.
In the top two rows of \cref{fig:supp-sampling}, we show $20$ visualizations of these base shapes randomly selected out of the $60$ value tokens in total.
We can also fuse these base shapes by linearly fusing the value tokens $\phi^\text{val}_k$ with a set of random weights (with a sum of $1$), and generate the a wide variety of animal shapes, as shown in the bottom two rows.

\begin{figure}[ht]
    \centering
    \includegraphics[width=\linewidth]{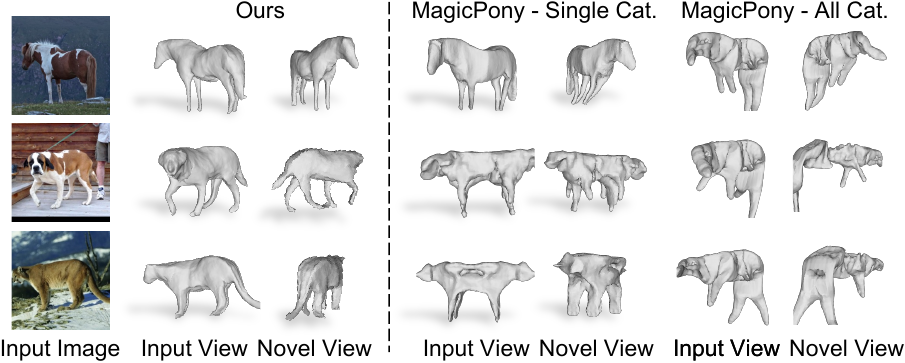}
    \caption{\textbf{Qualitative Comparisons} against two variants of MagicPony~\cite{wu2023magicpony}.
    In the middle are reconstruction results of the category-specific MagicPony model trained on individual categories.
    On the right are results of MagicPony trained on all categories jointly, \ie assuming all quadrupeds belong to one single category.
    }
    \vspace{-0.1in}
    \label{fig:supp-compare}
\end{figure}

\begin{figure*}[t]
    \centering
    \includegraphics[width=\linewidth]{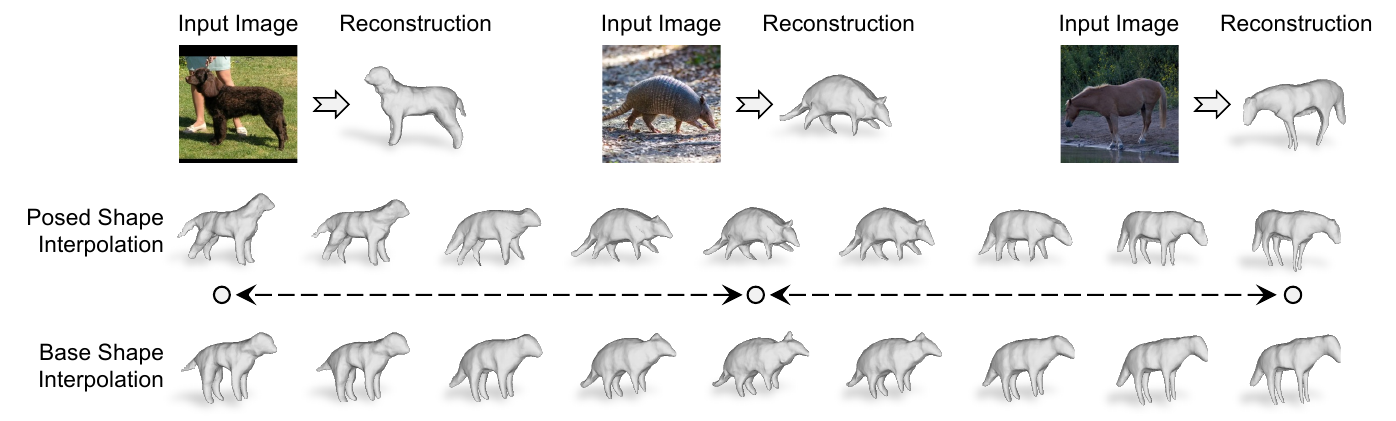}
    \caption{\textbf{Shape Interpolation between Instances.}
    On the top row, we show the 3D reconstructions from three input images.
    On the second and the third rows, we show the interpolation between the posed shapes and the base shapes.
    }
    \label{fig:supp-interpolation}
\end{figure*}
\begin{figure*}[t]
    \centering
    \includegraphics[width=\linewidth]{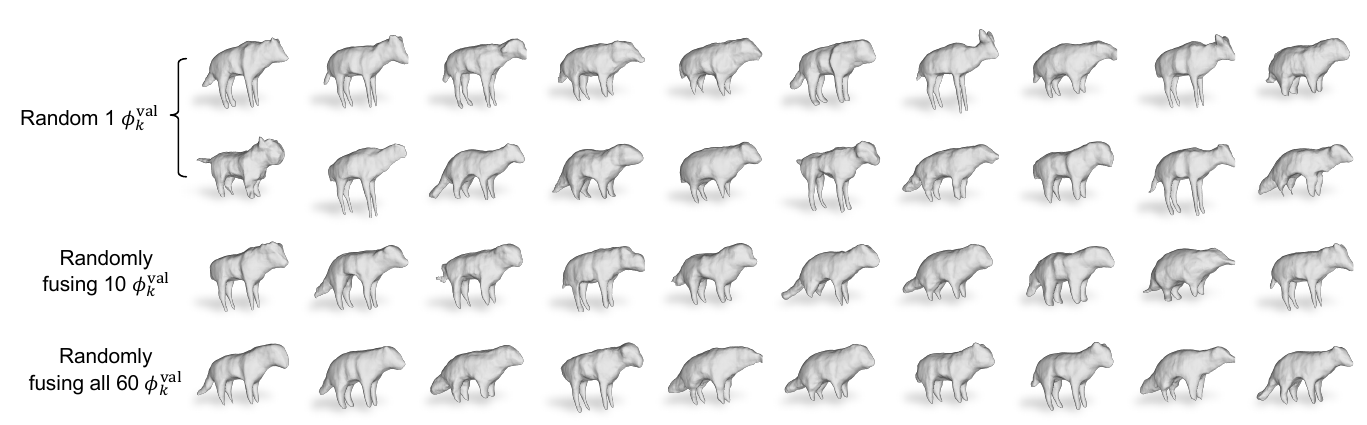}
    \caption{\textbf{Shape Generation from the Learned Semantic Bank.}
    On the top two rows, we visualize $20$ base shapes generated from the individual value tokens $\phi^\text{val}_k$ in the learned Semantic Bank.
    On the bottom two rows, we show the base shapes obtained by randomly fusing $10$ and $60$ value tokens $\phi^\text{val}_k$.
    }
    \label{fig:supp-sampling}
\end{figure*}

\subsection{Comparisons with Prior Work}

\paragraph{Quantitative Results for Each Category.}
\ifarxiv
Here, we provide the per-category performance break for the quantitative comparisons in \cref{tab:apt-animal3d-category}, which correspond to the aggregated results in \cref{tab:all}.
\else
Here, we provide the per-category performance break for the quantitative comparisons in \cref{tab:apt-animal3d-category}, which correspond to the aggregated results in Tab.~1 of the main paper.
\fi
On APT36K~\cite{yang2022apt}, we evaluate on four categories including horse, giraffe, cow and zebra.
On Animal3D~\cite{xu2023animal3d}, we use the available three categories: horse, cow and zebra.
Our pan-category model consistently outperforms the MagicPony~\cite{wu2023magicpony} baseline across all the categories, which highlights the benefits of the joint training of all categories. We also compare to LASSIE~\cite{yao2022lassie} and Hi-LASSIE~\cite{yao2023hi} quantitatively by optimizing on three Animal3D categories individually, as each category contains a small size ($< 100$) of images similar to the default setup proposed in their papers.

\begin{table}[t]
\small
\begin{minipage}[!t]{\columnwidth}
	\centering
    \begin{tabular}{lcccc}
\toprule
          & \multicolumn{4}{c}{APT-36K}                                        \\ \cmidrule(lr){2-5}
          & Horse          & Giraffe        & Cow            & Zebra          \\ \midrule
MagicPony~\cite{wu23magicpony} & 0.775          & 0.699          & 0.769          & 0.778          \\
Ours      & \textbf{0.853} & \textbf{0.796} & \textbf{0.876} & \textbf{0.840}  \\ \bottomrule
\end{tabular}
\end{minipage}
\\[10pt]
\begin{minipage}[!t]{\columnwidth}
	\centering
    \begin{tabular}{lccc}
\toprule
          & \multicolumn{3}{c}{Animal3D}                     \\
          \cmidrule(lr){2-4}
          & Horse          & Cow            & Zebra          \\
          \midrule
LASSIE~\cite{yao2022lassie}    & 0.850          & 0.887          & 0.878          \\
Hi-LASSIE~\cite{yao2023hi} & 0.410          & 0.720          & 0.704          \\
MagicPony~\cite{wu23magicpony} & 0.835          & 0.895          & 0.919          \\
Ours      & \textbf{0.884} & \textbf{0.903} & \textbf{0.942} \\
\bottomrule
\end{tabular}
\end{minipage}

\caption{
\textbf{Quantitative Comparisons} on APT-36K~\cite{yang2022apt} and Animal3D~\cite{xu2023animal3d} for each category. Our method consistently performs better than MagicPony~\cite{wu2023magicpony}, LASSIE~\cite{yao2022lassie} and Hi-LASSIE~\cite{yao2023hi} on all the categories.
}
\label{tab:apt-animal3d-category}
\end{table}

\begin{table}[t]
\small
\begin{minipage}[!t]{\columnwidth}
	\centering
    \begin{tabular}{lcccc}
\toprule
                     & \multicolumn{4}{c}{APT-36K}                                                                                    \\ \cmidrule(lr){2-5} 
                     & Horse                     & Giraffe                   & Cow                       & Zebra                     \\ \midrule
Final Model                 & \textbf{0.853}            & \textbf{0.796}            & \textbf{0.876}            & \textbf{0.840}            \\
w/o Semantic Bank    & 0.402                     & 0.398                     & 0.371                     & 0.373                     \\
Category-conditioned & \multicolumn{1}{l}{0.822} & \multicolumn{1}{l}{0.776} & \multicolumn{1}{l}{0.832} & \multicolumn{1}{l}{0.798} \\
w/o $\mathcal{L}_{\text{adv}}$      & \multicolumn{1}{l}{0.831} & \multicolumn{1}{l}{0.782} & \multicolumn{1}{l}{0.823} & \multicolumn{1}{l}{0.828} \\
\bottomrule
\end{tabular}
\end{minipage}
\\[10pt]
\begin{minipage}[!t]{\columnwidth}
	\centering
    \begin{tabular}{lccc}
\toprule
                     & \multicolumn{3}{c}{Animal3D}                                                      \\ \cmidrule(lr){2-4} 
                     & Horse                     & Cow                       & Zebra                     \\ \midrule
Final Model                 & \textbf{0.884}            & \textbf{0.903}            & \textbf{0.942}            \\
w/o Semantic Bank    & 0.402                     & 0.701                     & 0.630                     \\
Category-conditioned & \multicolumn{1}{l}{0.842} & \multicolumn{1}{l}{0.886} & \multicolumn{1}{l}{0.910} \\
w/o $\mathcal{L}_{\text{adv}}$      & \multicolumn{1}{l}{0.813} & \multicolumn{1}{l}{0.871} & \multicolumn{1}{l}{0.873} \\
\bottomrule
\end{tabular}
\end{minipage}

\caption{
\textbf{Quantitative Ablation Studies} on APT-36K~\cite{yang2022apt} and Animal3D~\cite{xu2023animal3d} for each category.
}
\label{tab:apt-animal3d-ablation}
\end{table}

\begin{table}[t]
\footnotesize
\begin{minipage}[!t]{\columnwidth}
    \setlength{\tabcolsep}{0.3cm}
	\centering
    \begin{tabular}{lccccc}
\toprule
$K$ & 2     & 10    & 60    & 100   & 500   \\ \midrule
PCK\@0.1       & 0.724 & 0.766 & 0.782 & 0.788 & 0.789 \\ \bottomrule
\end{tabular}
\end{minipage}
\caption{
\textbf{Bank Size Ablation Studies} on PASCAL~\cite{everingham2015pascal}.
}
\label{tab:bank-ablation}
\end{table}

\paragraph{MagicPony on All Categories.}

\ifarxiv
In \cref{fig:exp-compare-results}, 
\else
In Fig.~5 of the main paper,
\fi
we show that MagicPony~\cite{wu2023magicpony} fail to produce plausible 3D shapes when trained in a \emph{category-specific} fashion on species with limited ($<100$) number of images.
Alternatively, we can also train the MagicPony on our entire image dataset of all the animal species,
\ie treating all the images as in one single category.
The results are shown in \cref{fig:supp-compare}.
As MagicPony maintains only one single base shape for all animal instances, which is not able to capture the wide variation of shapes of different animal species.
On the contrary, our proposed Semantic Base Shape Bank learns various base shapes automatically adapted to different species, based on self-supervised image features.

\subsection{Quantitative Ablation Studies}
\ifarxiv
In addition to the qualitative comparisons in \cref{fig:ablation},
\else
In addition to the qualitative comparisons in Fig.~6 of the main paper,
\fi
\cref{tab:apt-animal3d-ablation} shows the quantitative ablation studies on APT-36K~\cite{yang2022apt} and Animal3D~\cite{xu2023animal3d}.
As explained in Sec.~5.3 of the paper,
we follow CMR~\cite{kanazawa2018learning} and optimize a linear mapping from our predicted vertices to the annotated keypoints in the \emph{input view}.
\ifarxiv
These numerical results are consistent with the visual comparisons in \cref{fig:ablation}.
\else
These numerical results are consistent with the visual comparisons in Fig.~6 of the main paper.
\fi

We also conducted additional experiments with different bank sizes, including $K = 2$, $10$, $60$, $100$, $500$, and report the PCK scores on PASCAL~\cite{everingham2015pascal} in \cref{tab:bank-ablation}. The quality grows with $K$; we pick $K = 60$ as a good trade-off with the computational cost.

\subsection{More Visualizations from \method}

We show more visualization results of \method on a wide variety of animals in \Cref{fig:supp-main-1}, \Cref{fig:supp-main-2} and \Cref{fig:supp-main-3}, including horse, weasel, pika, koala and so on.
Note that our model produces these articulated 3D reconstructions from just a single test image in feed-forward manner, without even knowing the category labels of the animal species.
With the articulated pose prediction, we can also easily animate the reconstructions in 3D.
More visualizations are presented at {\footnotesize\url{https://kyleleey.github.io/3DFauna/}}.

\subsection{Failure Cases and Limitations}

Despite promising results on a wide variety of quadruped animals, we still recognize a few limitations of the current method.
First, we only focus on quadrupeds which share a similar skeletal structure.
Although this covers a large number animals, including most mammals as well as many reptiles, amphibians and insects, the same assumption will not hold for many other animals in nature.
Jointly estimating the skeletal structure and 3D shapes directly from raw images remains a fundamental challenge for modeling the entire biodiversity.
Furthermore, for some fluffy animals that are highly deformable, like cats and squirrels, our model still struggles to reconstruct accurate poses and 3D shapes, as shown in \cref{fig:supp-failure}.

\begin{figure}[t]
    \centering
    \includegraphics[width=\linewidth]{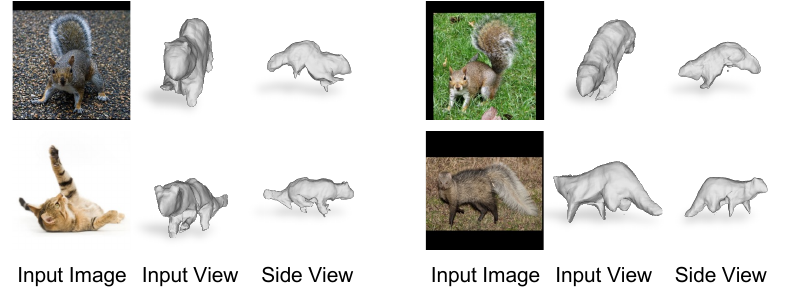}
    \caption{\textbf{Failure Cases.}
    For fluffy and highly deformable animals in challenging poses, our model still struggles in predicting the accurate poses and shapes.
    }
    \label{fig:supp-failure}
\end{figure}

Another failure case is the confusion of left and right legs, when reconstructing images taken from the side view,
for instance, in the second row of \cref{fig:supp-main-1}.
Since neither the object mask nor the self-supervised features~\cite{oquab2023dinov2} can provide sufficient signals to
disambiguate the legs, the model would ultimately have to resort to the subtle appearance cues, which still remains as a major challenge.
Finally, the current model still struggles at inferring high-fidelity appearance in a feed-forward manner, similar to \cite{wu2023magicpony}, and hence, we still employ a fast test-time optimization for better appearance reconstruction (within seconds).
This is partially due to the limited size of the dataset and the design of the texture field.
Leveraging powerful diffusion-based image generation models~\cite{siddiqui2022texturify} could provide additional signals to train a more effective 3D appearance predictor, which we plan to look into for future work.

\section{Additional Technical Details}

\subsection{Modeling Articulations}

In this work, we focus on quadruped animals which share a similar quadrupedal skeleton.
Here, we provide the details for the bone instantiation on the rest-pose shape based on a simple heuristic, the skinning model, and the additional bone rotation constraints.

\paragraph{Adaptive Bone Topology.}
We adopt a similar quadruped heuristic for rest-pose bone estimation as in \cite{wu2023magicpony}.
However, unlike \cite{wu2023magicpony} which focuses primarily on horses, our method needs to model a much more diverse set of animal species.
Hence, we make several modifications in order for the model to adapt to different animals automatically.
For the `spine', we still use a chain of 8 bones with equal lengths, connecting the center of the rest-pose mesh to the two most extreme vertices along $z$-axis.
To locate the four feet joints, we do not rely on the four $xz$-quadrants as the feet may not always land separately in those four quadrants, for instance, for animals with a longer body.
Instead, we locate the feet based on the distribution of the vertex locations.
Specifically, we first identify the vertices within the lower $40\%$ of the total height ($y$-axis).
We then use the center of these vertices as the origin of the $xz$-plane and locate the lowest vertex within each of the new quadrants as the feet joints.
For each leg, we create a chain of three bones of equally length connecting the foot joint to the nearest joint in the spine.

\paragraph{Bone Rotation Prediction.}
Similar to \cite{wu2023magicpony}, the viewpoint and bone rotations are predicted separately using different networks.
The viewpoint $\xi_1$ is predicted via a multi-hypothesis mechanism, as discussed in \cref{sec:sup-viewpoint}.
For the bone rotations $\xi_{2:B}$, we first project the middle point of each \emph{rest-pose} bone onto the image using the predicted viewpoint, and sample its corresponding local feature from the feature map using bilinear interpolation.
A Transformer-based~\cite{vaswani2017attention} network then fuses the global image feature, local image feature, 2D and 3D joint locations as well as the bone index, and produces the Euler angle for the rotation of each bone.
Unlike \cite{wu2023magicpony}, we empirically find it beneficial to add the bone index on top of other features instead of concatenation, which tends to encourage the model to separate the legs with different rotation predictions.

\paragraph{Skinning Weights.}
With the estimated bone structure, each bone $b$ except for the root has the parent bone $\pi(b)$.
Each vertex $V_{\text{ins},i}$ on the shape $V_{\text{ins}}$ is then associated to all the bones by skinning weights $w_{ib}$ defined as:
\begin{equation}
\begin{split}
    w_{ib} = \frac{e^{-d_{ib} / \tau_s}}{\sum_{k=1}^B e^{-d_{ik} / \tau_s}}, \quad \text{where} \\
    \quad d_{ib} = \mathop{\text{min}}\limits_{r\in[0,1]} ||V_{\text{ins},i} - r\tilde{\textbf{\text{J}}}_b - (1-r)\tilde{\textbf{\text{J}}}_{\pi(b)}||_2^2
\end{split}
\end{equation}
is the minimal distance from the vertex $V_{\text{ins},i}$ to each bone $b$, defined by the rest-pose joint location $\tilde{\textbf{\text{J}}}_b$ in world coordinates.
The $\tau_s$ is a temperature parameter set to $0.5$.
We then use the \emph{linear blend skinning equation} to pose the vertices:
\begin{equation}
    \begin{split}
        V_i(\xi) &= \left(\sum_{b=1}^B w_{ib}G_b(\xi)G_b(\xi^*)^{-1} \right)V_{\text{ins},i}, \\
        G_1 = g_1,\quad G_b &= G_{\pi(b)}\circ g_b,\quad g_b(\xi) = \begin{bmatrix}
            R_{\xi_b} & \textbf{\text{J}}_b \\ 0 & 1
        \end{bmatrix},
    \end{split}
\end{equation}
where the $\xi^*$ denotes the bone rotations at rest pose.

\paragraph{Bone Rotation Constraints.}

Following \cite{wu2023magicpony}, we regularize the magnitude of bone rotation predictions by $\cR_{\text{art}} = \frac{1}{B-1}\sum_{b=2}^B ||\xi_b||_2^2$.
In experiments, we find a common failure mode where instead of learning a reasonable shape with appropriate leg lengths, the model tends to predict excessively long legs for animals with shorter legs and bend them away from the camera.
To avoid this, we further constrain the range of the angle predictions.
Specifically, we forbid the rotation along $y$-axis (side-way) and $z$-axis (twist) of the lower two segments for each leg.
We also set a limit to the rotation along $y$-axis and $z$-axis of the upper segment for each leg as $(-10^\circ, 10^\circ)$.
For the body bones, we further limit the rotation along the $z$-axis within $(-6^\circ, 6^\circ)$.

\subsection{Viewpoint Learning Details}
\label{sec:sup-viewpoint}

Recovering the viewpoint of an object from only one input image is an ill-posed problem with numerous local optima in the reconstruction objective.
Here, we adopt the multi-hypothesis viewpoint prediction scheme introduced in~\cite{wu2023magicpony}.
In detail, our viewpoint prediction network outputs four viewpoint rotation hypotheses $R_k \in SO(3), k\in \{1,2,3,4\}$ within each of the four $xz$-quadrants together with their corresponding scores $\sigma_k$.
For computational efficiency, we randomly sample one hypothesis at each training iteration, and minimize the loss:
\begin{equation}
    \cL_{\text{hyp}}(\sigma_k, \cL_{\text{rec},k}) = (\sigma_k - \texttt{detach}(\cL_{\text{rec},k}))^2,
\end{equation}
where $\texttt{detach}$ indicates that the gradient on reconstruction loss is detached.
In this way, $\sigma_k$ essentially serves as an estimate of the expected reconstruction error for each hypothesis $k$, without actually evaluating it which would otherwise require the expensive rendering step.
During inference time, we can then take the $\texttt{softmax}$ of its inverse to obtain the probability $p_k$ of each hypothesis $k$: $p_k \propto \text{exp}(-\sigma_k / \tau)$, where the temperature parameter $\tau$ controls the sharpness of the distribution.

\subsection{Mask Discriminator Details}

To sample another viewpoint and render the mask for the mask discriminator, we randomly sample an azimuth angle and rotate the predicted viewpoint by that angle.
For conditioning, the detached input base embedding $\tilde{\phi}$ is concatenated to each pixel in the mask along the channel dimension, similar to CycleGAN~\cite{zhu2017unpaired}.
In practice, we also add a gradient penalty term in the discriminator loss following \cite{Niemeyer2020GIRAFFE,zhang2023seeing}.

\begin{table}[t]
\small
\begin{center}
\begin{tabular}{lc}
\toprule
 Parameter & Value/Range \\ \midrule
 Optimiser & Adam \\
 Learning rate on prior and bank & $1\times 10^{-3}$ \\
 Learning rate on others & $1\times 10^{-4}$ \\
 Number of iterations & $800$k \\
 Enable articulation iteration & $20$k \\
 Enable deformation iteration & $500$k \\
 Mask Discriminator iterations & $(80\text{k}, 300\text{k})$ \\
 Batch size & $6$ \\ \midrule
 Loss weight $\lambda_{\text{m}}$ & $10$ \\
 Loss weight $\lambda_{\text{im}}$ & $1$ \\
 Loss weight $\lambda_{\text{feat}}$ & $\{10, 1\}$ \\
 Loss weight $\lambda_{\text{Eik}}$ & $0.01$ \\
 Loss weight $\lambda_{\text{def}}$ & $10$ \\
 Loss weight $\lambda_{\text{art}}$ & $0.2$ \\
 Loss weight $\lambda_{\text{hyp}}$ & $\{50, 500\}$ \\ 
 Loss weight $\lambda_{\text{adv}}$ & $0.1$ \\ \midrule
 Image size & $256 \times 256$ \\
 Field of view (FOV) & $25^\circ$ \\
 Camera location & $(0, 0, 10)$ \\
 Tetrahedral grid size & $256$ \\
 Initial mesh centre & $(0, 0, 0)$ \\
 Translation in $x$- and $y$-axes & $(-0.4, 0.4)$ \\
 Translation in $z$-axis & $(-1.0, 1.0)$ \\
 Number of spine bones & $8$ \\
 Number of bones for each leg & $3$ \\
 Viewpoint hypothesis temperature $\tau$ & $(0.01, 1.0)$ \\
 Skinning weight temperature $\tau_\text{s}$ & $0.5$ \\
 Ambient light intensity $k_a$ & $(0.0, 1.0)$ \\
 Diffuse light intensity $k_d$ & $(0.5, 1.0)$ \\
\bottomrule
\end{tabular}
\end{center}
\caption{\textbf{Training details and hyper-parameter settings.}}\label{tab:sup-params}
\end{table}

\subsection{Network Architectures}

We adopt the architectures in \cite{wu2023magicpony} except the newly introduced Semantic Base Shape Bank and mask discriminator.
For the SBSM, we add a modulation layer~\cite{karras2020analyzing,karras2021alias} to each of the MLP layers to condition the SDF field on the base embeddings $\tilde{\phi}$.
To condition the DINO field, we simply concatenate the embedding to the input coordinates to the network.
The mask discriminator architecture is identical to that of
GIRAFFE~\cite{Niemeyer2020GIRAFFE}, except that we set input dimension as $129=1+128$, accommodating the $1$-channel mask and the $128$-channel shape embedding.
We set the size of the memory bank $K = 60$. 
In practice, to allow bank to represent categories with diverse kinds of shapes, we only fuse the value tokens with top $10$ cosine similarities.

\begin{figure}[t]
    \centering
    \includegraphics[width=\linewidth]{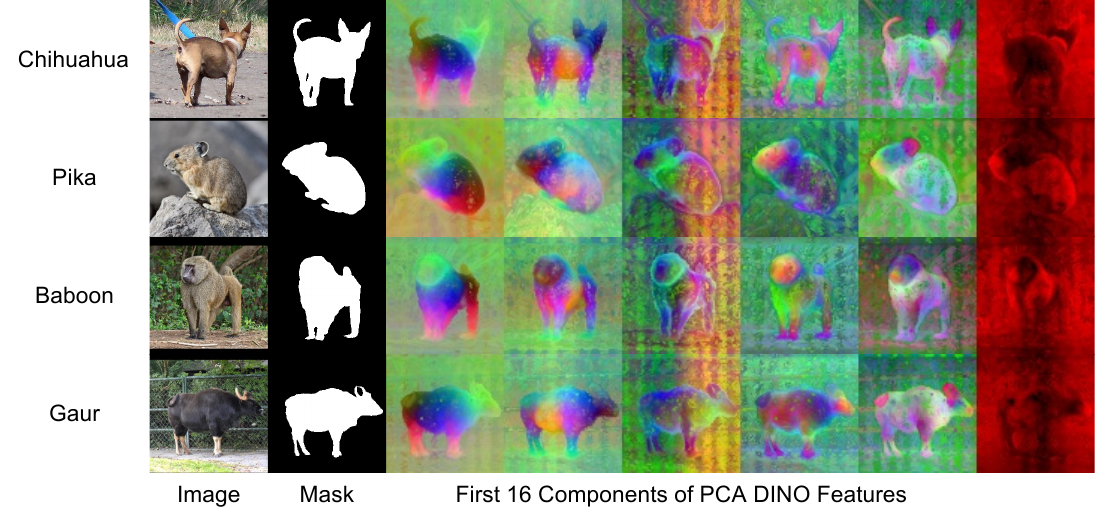}
    \caption{\textbf{Data Samples.}
    We show some samples of our training data. Each sample consists of the RGB image, automatically-obtained segmentation mask, and the corresponding $16$-channel PCA feature map.}
    \label{fig:data-sample}
\end{figure}

\subsection{Hyper-Parameters and Training Schedule}

The hyper-parameters and training details are listed in \cref{tab:sup-params}.
We train the model for $800$k iterations on a single NVIDIA A40 GPU, which takes roughly $5$ days.
In particular, we set $\lambda_{\text{feat}}$=10, and $\lambda_{\text{hyp}}$=50 at the start of training.
After $300$k iterations we change the values to $\lambda_{\text{feat}}$=1, $\lambda_{\text{hyp}}$=500.
During the first $6$k iterations, we allow the model to explore all four viewpoint hypotheses by randomly sampling
the four hypotheses uniformly, and gradually decrease the
chance of random sampling to $20\%$ while sampling the best
hypothesis for the rest $80\%$ of the time.
To save memory and computation, at each training iteration, we only feed images of the same species in a batch, and extract one base shape by averaging out the base embeddings.
At test time, we just directly use the shape embedding for each individual input image.

\subsection{Data Pre-Processing}

We use off-the-shelf segmentation models~\cite{kirillov2020pointrend,kirillov2023segment} to obtain the masks, crop around the objects and resize the crops to a size of $256 \times 256$. 
For the self-supervised features~\cite{oquab2023dinov2}, 
we randomly choose $5$k images from our dataset to compute the Principal Component Analysis~(PCA) matrix.
Then we use that matrix to run inference across all the images in our dataset.
We show some samples of different animal species in \cref{fig:data-sample}.
It is evident that these self-supervised image features can provide efficient semantic correspondences across different categories.
Note that masks are only for supervision, our model takes the raw image shown on the left as input for inference.

\subsection{Species Size Distribution}

\begin{figure}[t]
    \centering
    \includegraphics[width=\linewidth]{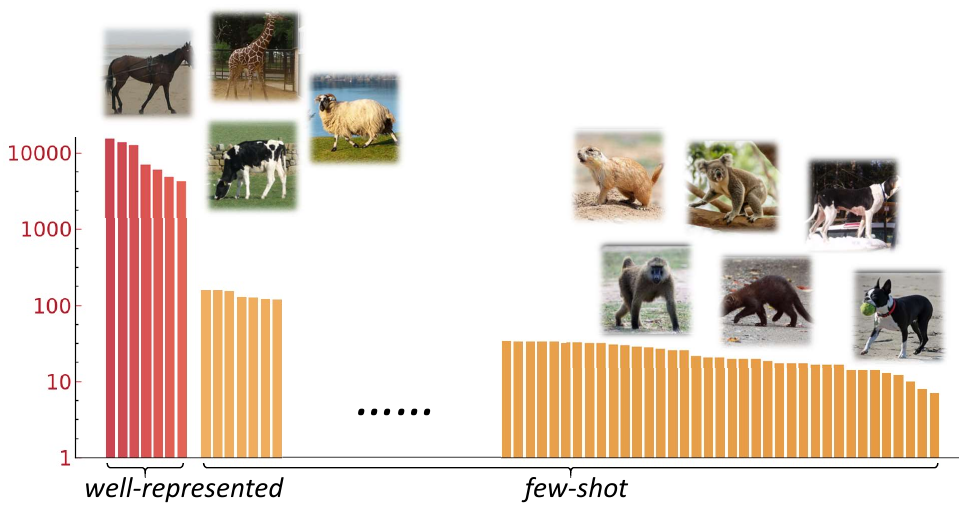}
    \caption{\textbf{Species Distribution.}
    We show the distribution of different animal species in our training dataset, including well-represented species with thousands of images and rare species with less than $100$ images.}
    \label{fig:data-static}
\end{figure}

We show a plot of the distribution of different species in our dataset below, including 7 well-represented categories~(\textcolor{red}{red}) and 121 few-shot categories~(\textcolor{orange}{orange}).
To balance the training, we duplicate the samples of few-shot categories to match the size of the rest.
\ifarxiv
Many examples in \cref{fig:exp-main-results} and \cref{fig:supp-main-1} in fact belong to the few-shot categories, such as koala, fisher and prairie dog.
\else
Many examples in Fig.~4 of the main paper and \cref{fig:supp-main-1} in fact belong to the few-shot categories, such as koala, fisher and prairie dog.
\fi

\begin{figure*}[t]
    \centering
    \includegraphics[width=0.96\linewidth]{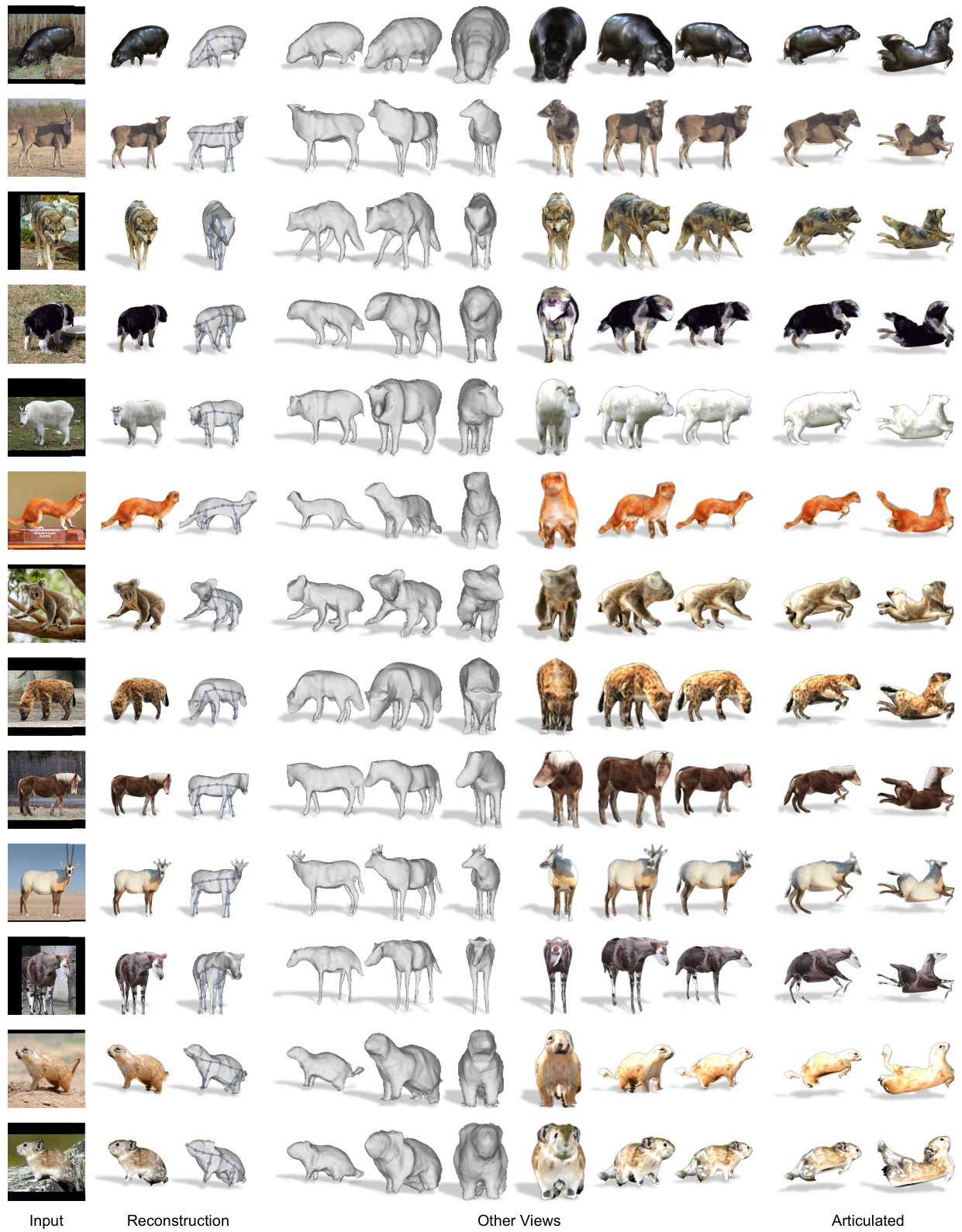}
    \caption{\textbf{Single Image 3D Reconstruction.}
    Given a single image of any quadruped animal at test time, our model reconstructs an articulated and textured 3D mesh in a feed-forward manner without requiring category labels, which can be readily animated.
    }
    \label{fig:supp-main-1}
\end{figure*}

\begin{figure*}[t]
    \centering
    \includegraphics[width=0.96\linewidth]{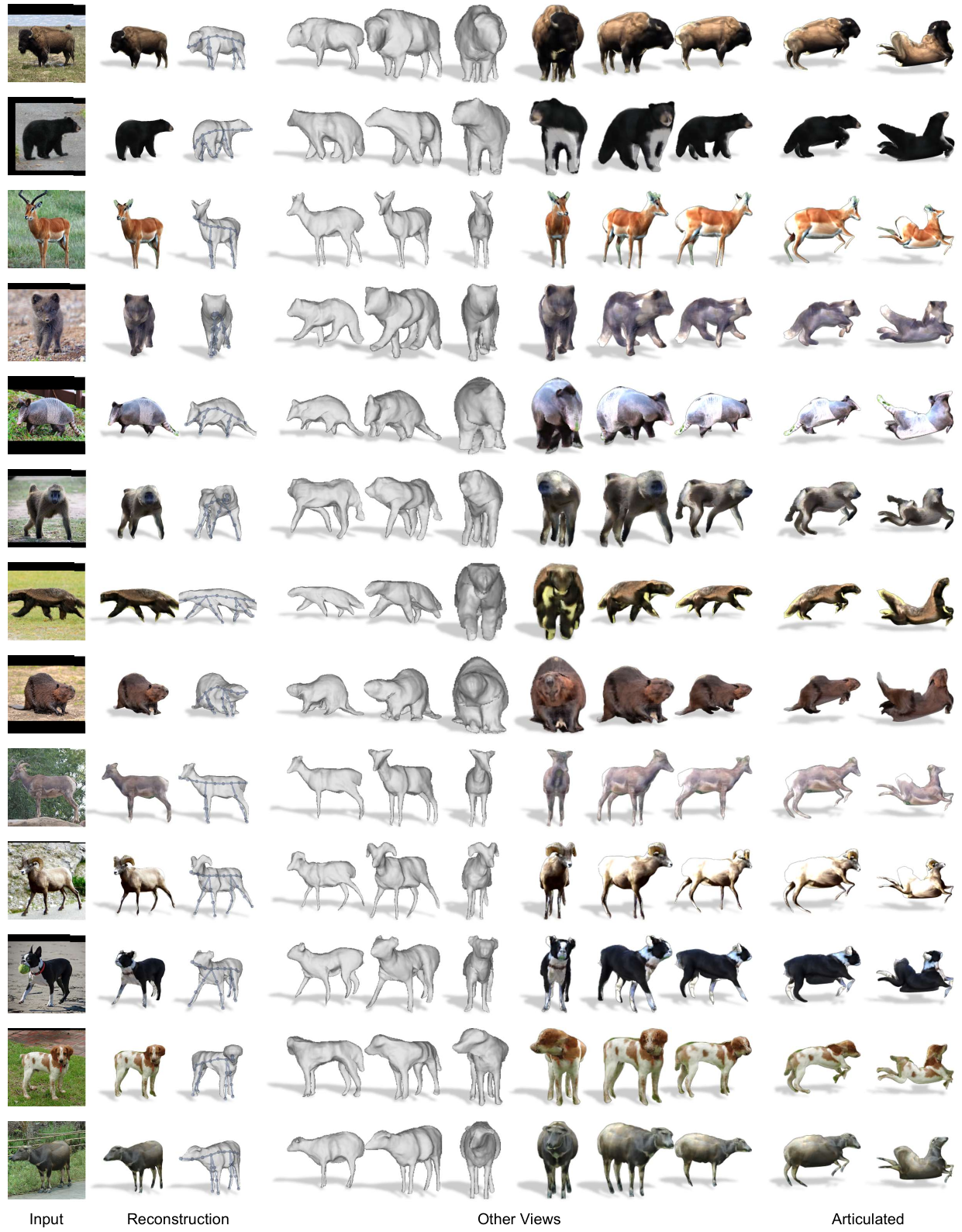}
    \caption{\textbf{Single Image 3D Reconstruction.}
    Given a single image of any quadruped animal at test time, our model reconstructs an articulated and textured 3D mesh in a feed-forward manner without requiring category labels, which can be readily animated.
    }
    \label{fig:supp-main-2}
\end{figure*}

\begin{figure*}[t]
    \centering
    \includegraphics[width=0.96\linewidth]{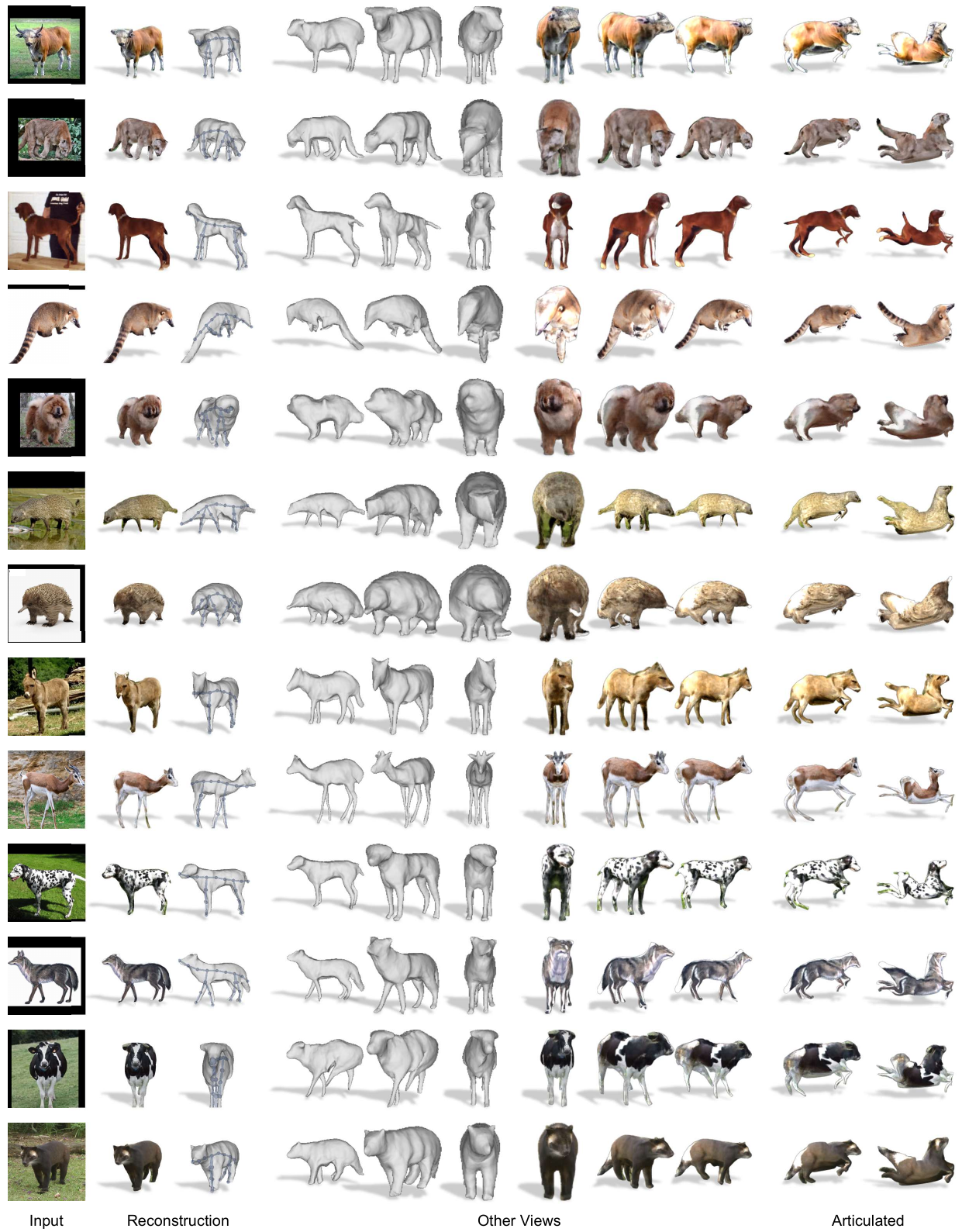}
    \caption{\textbf{Single Image 3D Reconstruction.}
    Given a single image of any quadruped animal at test time, our model reconstructs an articulated and textured 3D mesh in a feed-forward manner without requiring category labels, which can be readily animated.
    }
    \label{fig:supp-main-3}
\end{figure*}
\else
\fi

\end{document}